\documentclass[twoside,12pt]{article}
\usepackage{times,amsmath,amsthm,amsfonts,amsthm,amssymb,eucal}
\usepackage{graphicx,ifthen}
\usepackage{wrapfig}             
\usepackage{latexsym}
\usepackage{color}
\usepackage{mathrsfs}
\usepackage{bm}
\usepackage[square,numbers,sort]{natbib}
\usepackage{algorithm}
\usepackage{algorithmic}
\usepackage{caption}
\usepackage{subcaption}

\usepackage{tabularx}
\usepackage{array}
\usepackage{ragged2e}
\usepackage{lscape}
\usepackage{multirow}

\usepackage{url}

\begin{document}
\DeclareGraphicsExtensions{.gif,.pdf,.png,.jpg,.tiff}


\def\a{\alpha}
\def\I{{\bf I}}
\def\JJ{{\bf J}}
\def\mref#1{(\ref{#1})}
\def\NA{\text{NA}}
\def\g{\gamma_\NA}
\def\mRF#1{{$\text{mRF}_{\text{#1}}$}}
\def\nr{n_R}
\def\nl{n_L}
\def\tr{t_R}
\def\r{{\rho}}
\def\R{{\bf R}}
\def\rfo{{$\text{RF}_{\!\text{otf}\,\,}$}}
\def\rfoOne{{$\text{RF}_{\!\text{otf.1\,\,}}$}}
\def\rfoFive{{$\text{RF}_{\!\text{otf.5\,\,}}$}}
\def\rfor{{$\text{RF}_{\!\text{otfR\,\,}}$}}
\def\rforOne{{$\text{RF}_{\!\text{otfR.1\,\,}}$}}
\def\rforFive{{$\text{RF}_{\!\text{otfR.5\,\,}}$}}
\def\rfp{{$\text{RF}_{\!\text{prx}\,\,}$}}
\def\rfpk{{$\text{RF}_{\!\text{prx.k}\,\,}$}}
\def\rfpOne{{$\text{RF}_{\!\text{prx.1\,\,}}$}}
\def\rfpFive{{$\text{RF}_{\!\text{prx.5\,\,}}$}}
\def\rfpr{{$\text{RF}_{\!\text{prxR}\,\,}$}}
\def\rfprk{{$\text{RF}_{\!\text{prxR.k}\,\,}$}}
\def\rfprOne{{$\text{RF}_{\!\text{prxR.1\,\,}}$}}
\def\rfprFive{{$\text{RF}_{\!\text{prxR.5\,\,}}$}}
\def\rfu{{$\text{RF}_{\!\text{unsv}\,\,}$}}
\def\rfuOne{{$\text{RF}_{\!\text{unsv.1\,\,}}$}}
\def\rfuFive{{$\text{RF}_{\!\text{unsv.5\,\,}}$}}
\def\tl{t_L}
\def\trj{{t_{R_j}}}
\def\tlj{{t_{L_j}}}
\def\T{\Theta}
\def\V{\hat{\bf V}}
\def\X{{\bf X}}
\def\Xbar{{\overline X}}
\def\Y{{\bf Y}}
\def\ybar{\overline{y}}
\def\Ybar{\overline{Y}}
\def\Ymbar{\overline{\Y}}

\def\cc{\mathcal{C}}
\def\ee{\mathcal{E}}
\def\ii{\mathcal{I}}
\def\nn{\mathcal{N}}
\def\ss{\mathcal{S}}

\def\bct{\begin{center}}
\def\ect{\end{center}}
\def\BigHeading{\fontsize{15}{20}\bfseries}

\def\Array{\begin{eqnarray*}}
\def\EndArray{\end{eqnarray*}}
\def\Eq{\begin{equation}}
\def\EndEq{\end{equation}}

\def\Quote{\begin{quotation}\normalfont\small}
\def\EndQuote{\end{quotation}\rm}

\def\({\left(}
\def\){\right)}
\def\[{\left[}
\def\]{\right]}

\def\bct{\begin{center}}
\def\ect{\end{center}}
\def\Array{\begin{eqnarray*}}
\def\EndArray{\end{eqnarray*}}
\def\Enumerate{\begin{enumerate}}
\def\EndEnumerate{\end{enumerate}}
\def\Eq{\begin{equation}}
\def\EndEq{\end{equation}}
\def\EqArray{\begin{eqnarray}}
\def\EndEqArray{\end{eqnarray}}
\def\Itemize{\begin{itemize}}
\def\EndItemize{\end{itemize}}
\def\mref#1{(\ref{#1})}
\def\qt#1{\qquad\text{#1}}

\def\Demo#1{\par\ifdim\lastskip<\proclaimskipamount
            \removelastskip\proclaimskip\fi
        \sl#1. \hskip\smallindent\rm}
\def\EndDemo{\par\demoskip}
\def\DemoSection#1{\par\ifdim\lastskip<\proclaimskipamount
             \removelastskip\proclaimskip\fi
             #1\hskip\smallindent\rm}
\def\Section#1{\stepcounter{section}
    \DemoSection{{\bfseries\large\thesection.\hskip\smallindent#1}}}
\def\Subsection#1{\stepcounter{subsection}
    \DemoSection{\bfseries\normalsize\thesubsection.\hskip\smallindent#1}}
\def\Quote{\begin{quotation}\normalfont\small}
\def\EndQuote{\end{quotation}\rm}

\newtheorem{theorem}{Theorem}
\newtheorem{lemma}{Lemma}
\newtheorem{corollary}{Corollary}
\newtheorem{remark}{Remark}
\newtheorem{example}{Example}
\newtheorem{definition}{Definition}
\def\Theorem{\begin{theorem}\sl}
\def\EndTheorem{\end{theorem}}
\def\Lemma{\begin{lemma}\sl}
\def\EndLemma{\end{lemma}}
\def\Corollary{\begin{corollary}\sl}
\def\EndCorollary{\end{corollary}}
\def\Remark{\begin{remark}\rm}
\def\EndRemark{\end{remark}}
\def\Definition{\begin{definition}\sl}
\def\EndDefinition{\end{definition}}


\def\Report{Random Forest Missing Data Algorithms}
\def\Author{Tang and Ishwaran}
\pagestyle{myheadings}\markboth{\Report}{\Author}
\thispagestyle{empty}

\bct
{\BigHeading Random Forest Missing Data Algorithms}
\vskip10pt
Fei Tang and Hemant Ishwaran\\

\textit{Division of Biostatistics, University of Miami}

\vskip10pt
\rm\today 
\ect

\Quote\noindent
 Random forest (RF) missing data algorithms are an attractive approach
 for dealing with missing data.  They have the desirable properties of
 being able to handle mixed types of missing data, they are adaptive
 to interactions and nonlinearity, and they have the potential to
 scale to big data settings.  Currently there are many different RF
 imputation algorithms but relatively little guidance about their
 efficacy, which motivated us to study their performance.  Using a
 large, diverse collection of data sets, performance of various RF
 algorithms was assessed under different missing data mechanisms.
 Algorithms included proximity imputation, on the fly imputation, and
 imputation utilizing multivariate unsupervised and supervised
 splitting---the latter class representing a generalization of a new
 promising imputation algorithm called missForest.  
Performance of algorithms was
 assessed by ability to impute data accurately. Our findings
 reveal RF imputation to be generally robust with performance
 improving with increasing correlation.  Performance was good
 under moderate to high missingness, and even (in certain cases)
 when data was missing not at random.   
\EndQuote

\noindent
\rm Keywords:
{Imputation, Correlation, Missingness, Splitting (random,
univariate, multivariate, unsupervised), Machine Learning.}

\section{Introduction}

Missing data is a real world problem often encountered in scientific
settings.  Data that is missing is problematic as many statistical
analyses require complete data.  This forces researchers who want to
use a statistical analysis that requires complete data to choose
between imputing data or discarding missing values.  But to simply
discard missing data is not a reasonable practice, as valuable
information may be lost and inferential power
compromised~\cite{Enders:2010}. Thus, imputing missing data in such
settings is a more reasonable and practical way to proceed.

While many statistical methods have been developed for imputing
missing data, many of these perform poorly in high dimensional and
large scale data settings; for example, genomic, proteomic, neuroimaging,
and other high-throughput problems.  In particular, it is generally
recommended that all variables be included in multiple imputation to
make it proper in general and in order to not create bias in the
estimate of the correlations~\citep{Rubin:1996}. But this can lead to
overparameterization when there are a large number of variables and
the sample size is moderate.  Computational issues may also arise.  An
example is the occurrence of non-convexity due to missing data when
maximizing the log-likelihood.  This creates problems for traditional
optimization methods such as the EM
algorithm~\citep{LohWainwright:2011}. Missing data methods are also
often designed only for continuous data~(for example, gene expression
data~\citep{Aittokallio:2009}), and for those applicable to mixed data
(i.e., data having both nominal and categorical variables),
implementation can often break down in challenging data
settings~\citep{Liao:2014}. Another concern is the inability to deal
with complex interactions and nonlinearity of variables.  Standard
multiple imputation approaches do not automatically incorporate
interaction effects, which leads to biased parameter estimates when
interactions are present~\citep{Doove:2014}. Although some techniques,
such as fully conditional specification of covariates can be used to
try to resolve this problem~\citep{Bartlett:2015}, these techniques
can be difficult and inefficient to implement in settings where
interactions are expected to be complicated.

For these reasons there has been much interest in using machine
learning methods for missing data imputation.  A promising approach
can be based on Breiman's random
forests~\citep{Breiman:2001} (abbreviated hereafter as RF). RF have
the desired characteristic that they: (1) handle mixed types of
missing data; (2) address interactions and nonlinearity; (3) scale to
high-dimensions while avoiding overfitting; and (4) yield measures of
variable importance useful for variable selection.  Currently there
are several different RF missing data algorithms.  This includes the
original RF proximity algorithm proposed by
Breiman~\citep{Breiman:2003} implemented in the {\ttfamily
  randomForest} R-package~\citep{Liaw:2002}. A different class of
algorithms are the ``on-the-fly-imputation'' algorithms implemented in
the {\ttfamily randomSurvivalForest} R-package~\citep{IKBL:2008}, which
allow data to be imputed while simultaneously growing a survival tree.
These algorithms have been unified within the {\ttfamily
  randomForestSRC} R-package (abbreviated as RF-SRC) to include not
only survival, but classification and regression among other
settings~\citep{rfsrc}. A third approach is missForest, a method
recently introduced in~\citep{missForest}.
Missforest takes a different approach by recasting the missing data
problem as a prediction problem.  Data is imputed by regressing each
variable in turn against all other variables and then predicting
missing data for the dependent variable using the fitted forest.
MissForest has been shown~\citep{Waljee:2013} to outperform well known
methods such as $k$-nearest neighbors~\citep{Troyanskaya:2001} and
parametric MICE~\citep{VanBuuren:2007} (multivariate imputation using
chained equation).

Given that RF meets all the characteristics for handling missing data,
it seems desirable to use RF for imputing data.  However, there have
been no studies looking at the comparative behavior between RF missing
data algorithms.  There have been comparisons of RF to other
procedures, for example~\citep{missForest, Waljee:2013}, and there
have been studies looking at the effectiveness of RF imputation when
combined with other methods (for instance, one study showed that
parameter estimates were less biased when using MICE with random
forest based imputation~\cite{shah.mice}).  We therefore sought to
determine among RF algorithms which performed best, and under what
types of settings.  To study this, we used a large empirical study
involving a diverse collection of 60 data sets.  Performance was
assessed by imputation accuracy and computational speed.  Different
missing data mechanisms (missing at random and not missing at random)
were used to assess robustness.  In addition to the RF missing data
algorithms described above, we also considered several new
algorithms, including a multivariate version of missForest, referred
to as mForest.  Despite the superior performance of missForest (a
finding confirmed in our experiments), the algorithm is
computationally expensive to implement in high-dimensions as a
separate forest must be calculated for each variable and the algorithm
run until convergence is achieved.  The mForest algorithm helps to
alleviate this problem by grouping variables and using multivariate
forests with each group used in turn as the set of dependent
variables.  This replaces $p$ regressions, where $p$ is the number of
variables, with $\approx 1/\a$ regressions, where $0<\a<1$ is a user
specified group fraction size.  Computational savings were found to be
substantial for mForest without overly compromising accuracy even for
relatively large $\a$.  Other RF algorithms studied included a new
multivariate unsupervised algorithm and algorithms utilizing random
splitting.

For future reference, we note that all forests constructed in the
manuscript followed the RF methodology of~\citep{Breiman:2001}. Trees
were grown using independently sampled bootstrap data. For univariate
regression, continuous variables were split using squared-error
splitting; categorical variables by the Gini index~\citep{CART}. More
general splitting rules, such as unsupervised and multivariate
splitting, were also employed, and are described later in the
manuscript.  Random feature selection was used, with {\ttfamily mtry}
variables selected at random prior to splitting a tree node, and trees
were grown as deeply as possible subject to the constraint of a lower
bound of {\ttfamily nodesize} unique data values within a terminal
node.  All RF missing data algorithms were implemented using the
{\ttfamily randomForestSRC} R-package~\citep{rfsrc}, which has been
extended to include a new {\ttfamily impute.rfsrc} function optimized
specifically for data imputation.  The RF-SRC package implements
openMP parallel processing, which allows for parallel processing on
user desktops as well as large scale computing clusters; thus greatly
reducing computational times.

\section{RF approaches to imputation}
Three general strategies have been used for RF
missing data imputation:  
\Enumerate
\setlength\itemsep{-3pt}
\item[(A)] Preimpute the data; grow the forest; update the original
  missing values using proximity of the data. Iterate
  for improved results.
\item[(B)] Simultaneously impute data while growing the forest;
  iterate for improved results.
\item[(C)] Preimpute the data;  grow a forest using in turn each variable
  that has missing values; predict the missing values using the grown
  forest.  Iterate for improved results.
\EndEnumerate
\noindent
Proximity imputation~\citep{Breiman:2003} uses strategy A,
on-the-fly-imputation~\citep{IKBL:2008} (OTFI) uses strategy B, and
missforest~\citep{missForest} uses strategy C.  Below we detail each of
these strategies and describe various algorithms which utilize
one of these three approaches.  These new algorithms take
advantage of new splitting rules, including random splitting,
unsupervised splitting, and multivariate splitting~\citep{rfsrc}.

\subsection{Strawman imputation}

We first begin by describing a ``strawman imputation'' which will be
used throughout as our baseline reference value.  While this algorithm
is rough, it is also very rapid, and for this reason it was also used
to initialize some of our RF procedures.  Strawman imputation is
defined as follows.  Missing values for continuous variables are
imputed using the median of non-missing values, and for missing
categorical variables, the most frequently occurring non-missing value
is used (ties are broken at random).

\subsection{Proximity imputation: \rfp and \rfpr}

Here we describe proximity imputation (strategy A).  In this procedure
the data is first roughly imputed using strawman imputation.  A RF
is fit using this imputed data.  Using the resulting forest, the
$n\times n$ symmetric proximity matrix ($n$ equals the sample size) is
determined where the $( i, j)$ entry records the inbag frequency that
case $i$ and $j$ share the same terminal node.  The proximity matrix
is used to impute the original missing values.  For continuous
variables, the proximity weighted average of non-missing data is used;
for categorical variables, the largest average proximity over
non-missing data is used.  The updated data are used to grow a new RF,
and the procedure is iterated.

We use \rfp to refer to proximity imputation as described above.
However, when implementing \rfp we use a slightly modified version
that makes use of random splitting in order to increase computational
speed.  In random splitting, {\ttfamily nsplit}, a non-zero positive
integer, is specified by the user. A maximum of {\ttfamily
  nspit}-split points are chosen randomly for each of the randomly
selected {\ttfamily mtry} splitting variables.  This is in contrast to
non-random (deterministic) splitting typically used by RF, where all
possible split points for each of the potential {\ttfamily mtry}
splitting variables are considered. The splitting rule is applied to
the \textit{nsplit} randomly selected split points and the tree node
is split on the variable with random split point yielding the best
value, as measured by the splitting criterion.  Random splitting
evaluates the splitting rule over a much smaller number of split
points and is therefore considerably faster than deterministic
splitting.

The limiting case of random splitting is pure random splitting.  The
tree node is split by selecting a variable and the split-point
completely at random---no splitting rule is applied; i.e.\ splitting
is completely non-adaptive to the data.  Pure random splitting is
generally the fastest type of random splitting.  We also apply \rfp
using pure random splitting; this algorithm is denoted by \rfpr\!.

As an extension to the above methods, we implement iterated versions
of \rfp and \rfpr\!.  To distinguish between the different algorithms,
we write \rfpk and \rfprk when they are iterated $k\ge 1$ times.
Thus, \rfpFive and \rfprFive indicates that the algorithms were
iterated 5 times, while \rfpOne and \rfprOne indicates that the
algorithms were not iterated.  However, as this latter notation is
somewhat cumbersome, for notational simplicity we will simply use \rfp
to denote \rfpOne and \rfpr for \rfprOne\!.

\subsection{On-the-fly-imputation (OTFI): \rfo and \rfor}

A disadvantage of the proximity approach is that OOB (out-of-bag)
estimates for prediction error are biased~\citep{Breiman:2003}.
Further, because prediction error is biased, so are other measures
based on it, such as variable importance.  The method is also awkward
to implement on test data with missing values.  The OTFI
method~\citep{IKBL:2008} (strategy B) was devised to address these
issues.  Specific details of OTFI can be found in~\citep{IKBL:2008,rfsrc},
but for convenience we summarize the key aspects of OTFI below:
\Enumerate
\setlength\itemsep{-2pt}
\item Only non-missing data is used to calculate the split-statistic
  for splitting a tree node.  
\item
When assigning left and right daughter node membership if the variable
used to split the node has missing data, missing data for that
variable is ``imputed'' by drawing a random value from the inbag
non-missing data.  
\item Following a node split, imputed data are reset to missing and
  the process is repeated until terminal nodes are reached.  Note that
  after terminal node assignment, imputed data are reset back to
  missing, just as was done for all nodes.
\item Missing data in terminal nodes are then
  imputed using OOB non-missing terminal node data from all the trees.
  For integer valued variables, a maximal class rule is used; a mean
  rule is used for continuous variables.
\EndEnumerate

It should be emphasized that the purpose of the ``imputed data'' in Step 2
is only to make it possible to assign cases to daughter
nodes---imputed data is not used to calculate the split-statistic, and
imputed data is only temporary and reset to missing after node
assignment.  Thus, at the completion of growing the forest, the
resulting forest contains missing values in its terminal nodes and no
imputation has been done up to this point.  Step 4 is added as a means
for imputing the data, but this step could be skipped if the goal is
to use the forest in analysis situations.  In particular, step 4 is
not required if the goal is to use the forest for prediction.  This
applies even when test data used for prediction has missing values.
In such a scenario, test data assignment works in the same way as in
step 2.  That is, for missing test values, values are imputed as in
step 2 using the original grow distribution from the training forest,
and the test case assigned to its daughter node.  Following this, the
missing test data is reset back to missing as in step 3, and the
process repeated.

This method of assigning cases with missing data, which is well suited
for forests, is in contrast to surrogate splitting utilized by
CART~\citep{CART}.  To assign a case having a missing value for the
variable used to split a node, CART uses the best surrogate split
among those variables not missing for the case.  This ensures every
case can be classified optimally, whether the case has missing values
or not.  However, while surrogate splitting works well for CART, the
method is not well suited for forests.  Computational burden is one
issue.  Finding a surrogate split is computationally expensive even
for one tree, let alone for a large number of trees.  Another concern
is that surrogate splitting works tangentially to random feature
selection used by forests.  In RF, variables used to split a node are
selected randomly, and as such they may be uncorrelated, and a
reasonable surrogate split may not exist. Another concern is that
surrogate splitting alters the interpretation of a variable, which
affects measures such as variable importance
measures~\citep{IKBL:2008}.

To denote the OTFI missing data algorithm, we will use the
abbreviation \rfo\!.  As in proximity imputation, to increase
computational speed, \rfo is implemented using {\ttfamily nsplit}
random splitting.  We also consider OTFI under pure random splitting
and denote this algorithm by \rfor.  Both algorithms are iterated in
our studies.  \rfo\!, \rfor will be used to denote a single iteration,
while \rfoFive\!, \rforFive denotes five iterations.  Note that when
OTFI algorithms are iterated, the terminal node imputation executed in
step 4 uses inbag data and not OOB data after the first cycle.  This
is because after the first cycle of the algorithm, no coherent OOB
sample exists.

\Remark 
As noted by one of our referees, missingness
incorporated in attributes (MIA) is another tree splitting method which
bypasses the need to impute data~\cite{twala:2008, twala:2010}.  Again,
this only applies if the user is interested in a forest analysis.
MIA accomplishes this by
treating missing values as a category which is incorporated into the
splitting rule.  Let $X$ be an ordered or numeric feature being used
to split a node.  The MIA splitting rule searches over all possible
split values $s$ of $X$ of the following form:
\Enumerate
\setlength\itemsep{-2pt}
\item[]
Split A: $\{X\le s \text{ or } X = \text{ missing}\}$ versus $\{X > s\}$.
\item[]
Split B: $\{X\le s\}$ versus $\{X > s \text{ or } X = \text{ missing}\}$.
\item[]
Split C: $\{X = \text{ missing}\}$ versus $\{X = \text{ not missing}\}$.
\EndEnumerate
Thus, like OTF splitting, one can see that MIA results in a forest ensemble
constructed without having to impute data.  
\EndRemark

\subsection{Unsupervised imputation: \rfu}

\rfu refers to OTFI using multivariate unsupervised splitting.  However
unlike the OTFI algorithm, \rfo\!, the \rfu algorithm is
unsupervised and assumes there is no response (outcome)
variable.  Instead a multivariate unsupervised splitting
rule~\citep{rfsrc} is implemented.  As in the original RF algorithm,
at each tree node $t$, a set of {\ttfamily mtry} variables are
selected as potential splitting variables.  However, for each of
these, as there is no outcome variable, a random set of {\ttfamily
  ytry} variables is selected and defined to be the multivariate
response (pseudo-outcomes).  A multivariate composite splitting rule
of dimension {\ttfamily ytry} (see below) is applied to each of the {\ttfamily
  mtry} multivariate regression problems and the node $t$ split on the
variable leading to the best split.  Missing values in the response
are excluded when computing the composite multivariate splitting rule:
the split-rule is averaged over non-missing responses
only~\citep{rfsrc}. We also consider an iterated \rfu algorithm (e.g.\
\rfuFive implies five iterations, \rfu implies no iterations).

Here is the description of the multivariate composite splitting rule.  We begin by
considering univariate regression.  For notational simplicity, let us
suppose the node $t$ we are splitting is the root node based on the full
sample size $n$. 
 Let $X$ be the feature used to split $t$, where for simplicity 
we assume $X$ is ordered or numeric.
Let $s$ be a proposed split for $X$ that splits $t$ into left and
right daughter nodes $\tl:=\tl(s)$ and $\tr:=\tr(s)$, where
$\tl=\{X_i\le s\}$ and $\tr=\{X_i>s\}$.  Let $\nl=\#\tl$ and
$\nr=\#\tr$ denote the sample sizes for the two daughter nodes.  If
$Y_i$ denotes the outcome, the squared-error split-statistic for the
proposed split is
$$
D(s,t)
=\frac{1}{n}\sum_{i\in \tl}(Y_i-\Ybar_{\tl})^2 
 + \frac{1}{n}\sum_{i\in \tr}(Y_i-\Ybar_{\tr})^2 
$$
where $\Ybar_{\tl}$ and $\Ybar_{\tr}$ are the sample means for
$\tl$ and $\tr$ respectively.  The best split for $X$ is the
split-point $s$ minimizing $D(s,t)$.  To extend the squared-error
splitting rule to the multivariate case $q>1$, we apply univariate
splitting to each response coordinate separately.  Let
$\Y_i=(Y_{i,1},\ldots,Y_{i,q})^T$ denote the $q\ge 1$ dimensional outcome.
For multivariate regression analysis, an averaged standardized
variance splitting rule is used.  The goal is to minimize
$$
D_q(s,t)
=\sum_{j=1}^q \left\{\sum_{i\in \tl}(Y_{i,j}-\Ybar_{\tlj})^2 
 + \sum_{i\in \tr}(Y_{i,j}-\Ybar_{\trj})^2\right\}
$$ 
where $\Ybar_{\tlj}$ and $\Ybar_{\trj}$ are the sample means of the
$j$-th response coordinate in the left and right daughter nodes.
Notice that such a splitting rule can only be effective if each of the
coordinates of the outcome are measured on the same scale, otherwise
we could have a coordinate $j$, with say very large values, and its
contribution would dominate $D_q(s,t)$.  We therefore calibrate
$D_q(s,t)$ by assuming that each coordinate has been standardized
according to
$$
\frac{1}{n}\sum_{i\in t} Y_{i,j} = 0,\hskip10pt
\frac{1}{n}\sum_{i\in t} Y_{i,j}^2 = 1,\hskip10pt
1\le j\le q.
\label{y.stand.assumption}
$$
The standardization is applied prior to splitting a node.  To make
this standardization clear, we denote the standardized responses by
$Y_{i,j}^*$.  With some elementary manipulations, it can be
verified that minimizing $D_q(s,t)$ is equivalent to maximizing
\Eq
D_q^*(s,t) = \sum_{j=1}^q \left\{ \frac{1}{\nl}\(\sum_{i\in
  \tl}Y_{i,j}^*\)^2 + \frac{1}{\nr}\(\sum_{i\in \tr}Y_{i,j}^*\)^2
\right\}.
\label{mvr.split}
\EndEq
For multivariate classification, an averaged standardized
Gini splitting rule is used.  First consider the univariate case
(i.e., the multiclass problem) where the outcome $Y_i$ is a class label
from the set $\{1,\ldots,K\}$ where $K\ge 2$.  The best split $s$ for $X$
is obtained by maximizing
$$
G(s,t) = 
\sum_{k=1}^K \[
\frac{1}{\nl}\(\sum_{i\in \tl} Z_{i(k)}\)^2 
+ \frac{1}{\nr}\(\sum_{i\in \tr} Z_{i(k)}\)^2 \]
$$ 
where $Z_{i(k)}=1_{\{Y_i=k\}}$.  Now consider the multivariate
classification scenario $r>1$, where each outcome coordinate $Y_{i,j}$
for $1\le j\le r$ is a class label from $\{1,\ldots,K_j\}$ for
$K_j\ge 2$.  We apply Gini splitting to each coordinate yielding the
extended Gini splitting rule
\Eq
G_r^*(s,t)  = 
\sum_{j=1}^r \[ \frac{1}{K_j}\sum_{k=1}^{K_j} \left\{
\frac{1}{\nl}\(\sum_{i\in \tl} Z_{i(k),j}\)^2 
+ \frac{1}{\nr}\(\sum_{i\in \tr} Z_{i(k),j}\)^2 
\right\}\]
\label{egini.split}
\EndEq
where $Z_{i(k),j}=1_{\{Y_{i,j}=k\}}$.  Note that
the normalization $1/K_j$ employed for a coordinate $j$ is
required to standardize the contribution of the Gini split from that
coordinate.

Observe that~\mref{mvr.split} and~\mref{egini.split} are equivalent
optimization problems, with optimization over $Y_{i,j}^*$ for
regression and $Z_{i(k),j}$ for classification.  As shown
in~\citep{Ishwaran:2015} this leads to similar theoretical splitting
properties in regression and classification settings.  Given this
equivalence, we can combine the two splitting rules to form a
composite splitting rule.  The mixed outcome splitting rule $\T(s,t)$
is a composite standardized split rule of mean-squared
error~\mref{mvr.split} and Gini index splitting~\mref{egini.split};
i.e.,
$$
\T(s,t) = D_q^*(s,t)+G_r^*(s,t),
$$ 
where $p=q+r$.  The best split for $X$ is the value of $s$
maximizing $\T(s,t)$.  

\Remark As discussed in~\citep{Segal:2011}, multivariate regression
splitting rules patterned after the Mahalanobis distance can be used
to incoporate correlation between response coordinates.  Let
$\Ymbar_L$ and $\Ymbar_R$ be the multivariate means for $\Y$ in
the left and right daughter nodes, respectively.  The following
Mahalanobis distance splitting rule was discussed
in~\citep{Segal:2011}
$$
M_q(s,t)
= \sum_{i\in \tl}\(\Y_i-\Ymbar_{L}\)^T\V_L^{-1}\(\Y_i-\Ymbar_{L}\)
 + \sum_{i\in \tr}\(\Y_i-\Ymbar_{R}\)^T\V_R^{-1}\(\Y_i-\Ymbar_{R}\)
$$ 
where $\V_L$ and $\V_R$ are the estimated covariance matrices for the
left and right daughter nodes.  While this is a reasonable approach in
low dimensional problems, recall that we are applying $D_q(s,t)$ to
{\ttfamily ytry} of the feature variables which could be large if the
feature space dimension $p$ is large.  Also, because of missing data
in the features, it may be difficult to derive estimators for $\V_L$
and $\V_R$, which is further complicated if their dimensions are high.
This problem becomes worse as the tree is grown because the number of
observations decreases rapidly making estimation unstable.  For these
reasons, we use the splitting rule $D_q(s,t)$ rather than $M_q(s,t)$
when implementing imputation.  \EndRemark

\subsection{mForest imputation: \mRF{$\a$} and mRF}

The missForest algorithm recasts the missing data problem as a
prediction problem.  Data is imputed by regressing each variable in
turn against all other variables and then predicting missing data for
the dependent variable using the fitted forest.  With $p$ variables,
this means that $p$ forests must be fit for each iteration, which can
be slow in certain problems.  Therefore, we introduce a
computationally faster version of missForest, which we call mForest.
The new algorithm is described as follows.  Do a quick strawman
imputation of the data.  The $p$ variables in the data set are
randomly assigned into mutually exclusive groups of approximate size
$\a p$ where $0<\a<1$.  Each group in turn acts as the multivariate
response to be regressed on the remaining variables (of approximate
size $(1-\a)p$).  Over the multivariate responses, set imputed values
back to missing.  Grow a forest using composite multivariate
splitting.  As in \rfu, missing values in the response are excluded
when using multivariate splitting: the split-rule is averaged over
non-missing responses only.  Upon completion of the forest, the
missing response values are imputed using prediction.  Cycle over all
of the $\approx 1/\a$ multivariate regressions in turn; thus
completing one iteration.  Check if the imputation accuracy of the
current imputed data relative to the previously imputed data has
increased beyond an $\epsilon$-tolerance value (see Section 3.3 for
measuring imputation accuracy).  Stop if it has, otherwise repeat
until convergence.

To distinguish between mForest under different $\a$ values we use
the notation \mRF{$\a$} to indicate mForest fit under the specified
$\a$.  Note that the limiting case $\a=1/p$ corresponds to
missForest.  Although technically the two algorithms missForest and
\mRF{$\a$} for $\a=1/p$ are slightly different, we did not find
significant differences between them during informal experimentation.
Therefore for computational speed, missForest was taken to be
\mRF{$\a$} for $\a=1/p$ and denoted simply as mRF.

\section{Methods}
\subsection{Experimental design and data}

Table~\ref{expt} lists the nine experiments that were carried out.  In each
experiment, a pre-specified target percentage of missing values was
induced using one of three different missing mechanisms~\citep{Rubin:1976}: 
\Enumerate
\setlength\itemsep{-2pt}
\item
Missing
completely at random~(MCAR).  This means the probability that an
observation is missing does not depend on the observed value or the
missing ones.
\item
Missing at random~(MAR).  This means that the probability of
missingness may depend upon observed values, but not missing ones.
\item
Not missing at random~(NMAR).  This means the probability of missingness
depends on both observed and missing values.
\EndEnumerate

\begin{table}[htp!]
\centering
\caption{Experimental design used for large scale study of RF missing data algorithms.}
\begin{tabular}{ccc}
  \hline
 & Missing   & Percent  \\ 
 & Mechanism & Missing  \\ 
  \hline
EXPT-A & MCAR  &25 \\ 
EXPT-B & MCAR  &50 \\ 
EXPT-C & MCAR  &75 \\ 
EXPT-D & MAR   &25 \\ 
EXPT-E & MAR   &50 \\ 
EXPT-F & MAR   &75 \\ 
EXPT-G & NMAR  &25 \\ 
EXPT-H & NMAR  &50 \\ 
EXPT-I & NMAR  &75 \\   \hline
\end{tabular}
\label{expt}
\end{table}

\begin{figure*}[htp!]
\begin{center}
\hskip-20pt
\resizebox{5.30in}{!}{\includegraphics[page=1]{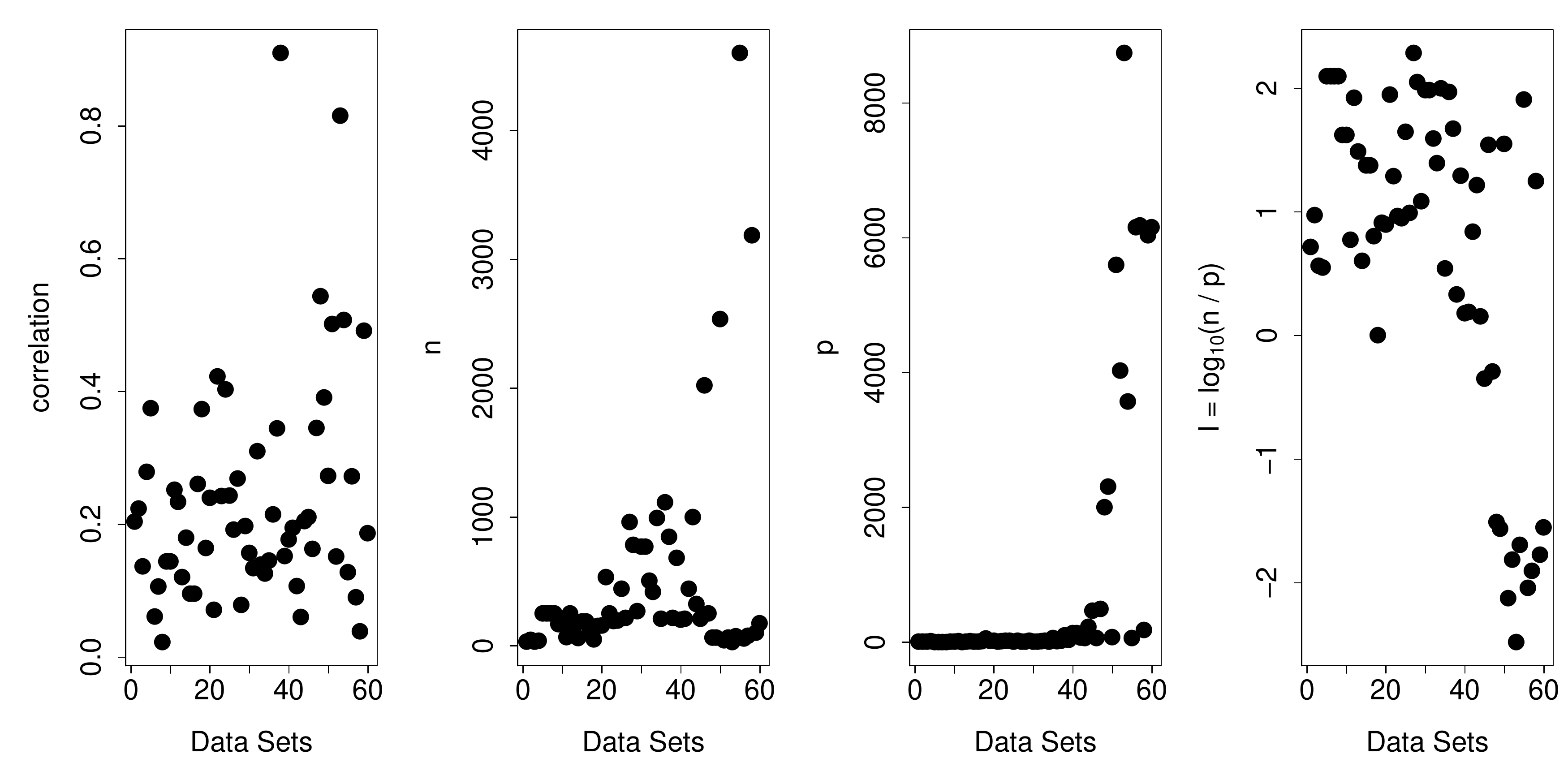}}
\end{center}
\vskip-15pt
\caption{Summary values for the 60 data sets used in the large scale
  RF missing data experiment.  The last panel displays the log-information,
  $I=\log_{10}(n/p)$, for each data set.}
\label{data}
\end{figure*}

Sixty data sets were used, including both real and synthetic data.
Figure~\ref{data} illustrates the diversity of the data.  Displayed
are data sets in terms of correlation ($\rho$), sample size ($n$),
number of variables ($p$), and the amount of information 
contained in the data ($I=\log_{10}(n/p)$).  The correlation, $\rho$, was defined as the
$L_2$-norm of the correlation matrix.  If $\R=(\rho_{i,j})$ denotes
the $p\times p$ correlation matrix for a data set, $\rho$ was defined
to equal 
\Eq
\rho=\binom{p}{2}^{-1}\sum_{j=1}^p\(\sum_{k<j}|\rho_{i,j}|^2\)^{1/2}.
\label{correlation}
\EndEq
This is similar to the usual definition of the $L_2$-norm for a
matrix, but where we have modifed the definition to remove the
diagonal elements of $\R$ which equal 1, as well as the contribution from the
symmetric lower diagonal values.

Note that in the plot for $p$ there are 10 data sets with $p$ in
the thousands---these are a collection of well known gene expression
data sets.  The right-most plot displays the log-information of a data
set, $I=\log_{10}(n/p)$.  The range of values on the log-scale vary
from $-2$ to 2; thus the information contained in a data set can
differ by as much as $\approx 10^4$.

\vskip10pt
\vskip-10pt

\subsection{Inducing MCAR, MAR and NMAR missingness}

The following procedures were used to induce missigness in the data.
Let the target missigness fraction be $0<\g<1$.  For MCAR, data was
set to missing randomly without imposing column or row constraints
to the data matrix.  Specifically, the data matrix was made into a
long vector and $n\g$ of the entries selected at random and set to
missing.

For MAR, missing values were assigned by column.  Let
$\X_j=(X_{1,j},\ldots,X_{n,j})$ be the $n$-dimensional vector
containing the original values of the $j$th variable, $1\le j\le p$.
Each coordinate of $\X_j$ was made missing according to the tail
behavior of a randomly selected covariate $\X_k$, where $k\ne j$.  The
probability of selecting coordinate $X_{i,j}$ was
$$
P\{\text{selecting } X_{i,j}|B_{j}\} \propto
\left\{ \begin{array}{cc}
F(X_{i,k})     &\text{if } B_{j}=1\\
1 - F(X_{i,k}) &\text{if } B_{j}=0\\
\end{array}\right.
$$
where $F(x)=(1+\exp(-3 x))^{-1}$ and $B_j$ were i.i.d.\ symmetric 0-1
Bernoulli random variables.  With this method, about half of the
variables will have higher missigness in those coordinates
corresponding to the right tail of a randomly selected variable (the
other half will have higher missigness depending on the left tail of a
randomly selected variable). A total of $n\g$ coordinates were
selected from $\X_j$ and set to missing.  This induces MAR, as missing
values for $\X_j$ depend only on observed values of another variable
$\X_k$.

For NMAR, each coordinate of $\X_j$ was made missing according to its
own tail behavior.  A total of $n\g$ values were selected according to
$$
P\{\text{selecting } X_{i,j}\} \propto
\left\{ \begin{array}{cc}
F(X_{i,j})     &\text{with probability 1/2}\\
1 - F(X_{i,j}) &\text{with probability 1/2}.
\end{array}\right.
$$
Notice that missingness in $X_{i,j}$ depends on both observed and missing
values.  In particular, missing values occur with higher probability
in the right and left tails of the empirical distribution.  
Therefore, this induces NMAR.

\subsection{Measuring imputation accuracy}

Accuracy of imputation was assessed using the following metric.  As
described above, values of $\X_j$ were made missing under various
missing data assumptions.  Let $(1_{1,j},\ldots,1_{n,j})$ be a
vector of zeroes and ones indicating which values of $\X_j$ were
artificially made missing.  Define $1_{i,j}=1$ if $X_{i,j}$ is
artificially missing; otherwise $1_{i,j}=0$.  Let
$n_j=\sum_{i=1}^n1_{i,j}$ be the number of artificially induced
missing values for $\X_j$.

Let $\nn$ and $\cc$ be the set of nominal (continuous) and categorical
(factor) variables with more than one artificially induced missing
value.  That is,
\Array
\nn&=& \{j:\X_j  \text{ is nominal and $n_j>1$}\}\\
\cc&=& \{j:\X_j  \text{ is categorical and $n_j>1$}\}.
\EndArray
Standardized root-mean-squared error (RMSE) was used to assess
performance for nominal variables, and misclassification error for
factors.  Let $\X^*_j$ be the $n$-dimensional vector of imputed values
for $\X_j$ using procedure $\ii$.  Imputation error for $\ii$
was measured using
\Array
\ee(\ii)
&=&
\frac{1}{\#\nn}\sum_{j\in\nn}\sqrt{
\frac{\displaystyle\sum_{i=1}^n 1_{i,j} \(X_{i,j}^* - X_{i,j} \)^2/n_j}
{\displaystyle\sum_{i=1}^n 1_{i,j} \(X_{i,j} - \Xbar_j\)^2/n_j}}\\
&&  \qquad + 
\frac{1}{\#\cc}\sum_{j\in\cc}\[
\dfrac{\sum_{i=1}^n 1_{i,j}\, 1\{X_{i,j}^* \ne X_{i,j}\}}
{n_j }\],
\EndArray 
where $\Xbar_j=\sum_{i=1}^n \(1_{i,j}X_{i,j}\)/n_j$.  To be clear
regarding the standardized RMSE, observe that the denominator in the
first term is the variance of $\X_j$ over the
artificially induced missing values, while the numerator is the MSE
difference of $\X_j$ and $\X_j^*$ over the induced missing values.

As a benchmark for assessing imputation accuracy we used strawman
imputation described earlier, which we denote by $\ss$.  Imputation
error for a procedure $\ii$ was compared to $\ss$ using relative
imputation error defined as
$$
\ee_R(\ii) = 100 \times \dfrac{\ee(\ii)}{\ee(\ss)}.
$$
A value of less than 100 indicates a procedure $\ii$ performing better
than the strawman.

\subsection{Experimental settings for procedures}

Randomized splitting was invoked with an {\ttfamily nsplit} value of
10.  For random feature selection, {\ttfamily mtry} was set to
$\sqrt{p}$.  For random outcome selection for \rfu\!, we set {\ttfamily
  ytry} to equal $\sqrt{p}$.  Algorithms \rfo, \rfu and \rfp were
iterated 5 times in addition to be run for a single iteration.
For mForest, the percentage of variables used as responses was
$\a=.05, .25$.  This implies that \mRF{0.05} used up to 20 regressions
per cycle, while \mRF{0.25} used 4.  Forests for all procedures were
grown using a {\ttfamily nodesize} value of 1.  Number of trees was
set at {\ttfamily ntree} $=500$.  Each experimental setting
(Table~\ref{expt}) was run 100 times independently and results
averaged.

For comparison, $k$-nearest neighbors imputation (hereafter denoted as
KNN) was applied using the {\ttfamily impute.knn} function from the
R-package {\ttfamily impute}~\citep{impute}.  For each data point with
missing values, the algorithm determines the $k$-nearest neighbors
using a Euclidean metric, confined to the columns for which that data
point is not missing.  The missing elements for the data point are
then imputed by averaging the non-missing elements of its neighbors.
The number of neighbors $k$ was set at the default value $k=10$.  In
experimentation we found the method robust to the value of $k$ and
therefore opted to use the default setting.  Much more important were
the parameters {\ttfamily rowmax} and {\ttfamily colmax} which control
the maximum percent missing data allowed in each row and column of the
data matrix before a rough overall mean is used to impute the
row/column.  The default values of 0.5 and 0.8, respectively, were too
low and led to poor performance in the heavy missing data experiments.
Therefore, these values were set to their maximum of 1.0, which
greatly improved performance.  Our rationale for selecting KNN as a
comparison procedure is due to its speed because of the large scale
nature of experiments (total of $100\times 60 \times 9 =54,000$ runs
for each method).  Another reason was because of its close
relationship to forests.  This is because RF is also a type of nearest
neighbor procedure---although it is an adaptive nearest neighbor.  We
comment later on how adaptivity may give RF a distinct advantage over
KNN.  

\section{Results}

Section 4.1 presents the results of the performance of a procedure as
measured by relative imputation accuracy, $\ee_R(\ii)$, and in Section
4.2 we discuss computational speed.

\subsection{Imputation Accuracy}
\begin{figure*}[htp!]
\begin{center}
\vskip10pt
\resizebox{5.25in}{!}{\includegraphics[page=1]{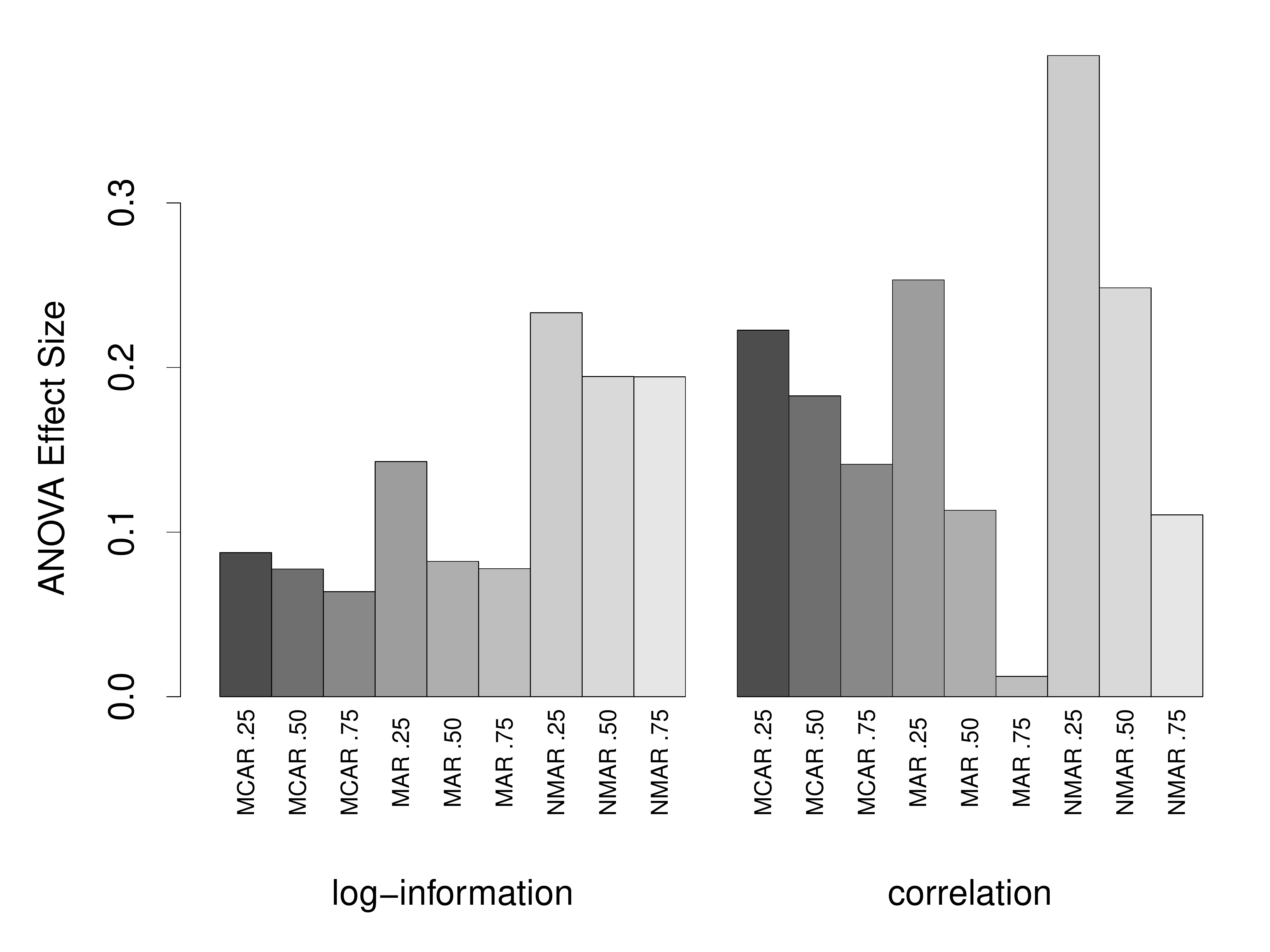}}
\end{center}
\vskip-10pt
\caption{ANOVA effect size for the log-information,
  $I=\log_{10}(n/p)$, and correlation, $\rho$ (defined as
  in~\mref{correlation}), from a linear regression using log relative
  imputation error, $\log_{10}(\ee_R(\ii))$, as the response.  In addition
  to $I$ and $\rho$, dependent variables in the regression included
  type of RF procedure used.  ANOVA effect sizes are the estimated
  coefficients of the standardized variable (standardized to have mean
  zero and variance 1).}
\label{anova}
\end{figure*}

In reporting the values for imputation accuracy, we have stratified
data sets into low, medium and high-correlation groups, where
correlation, $\r$, was defined as in~\mref{correlation}.  Low, medium
and high-correlation groups were defined as groups whose $\rho$ value
fell into the $[0,50]$, $[50,75]$ and $[75, 100]$ percentile for
correlations.  Results were stratified by $\rho$ because we found it
played a very heavy role in imputation performance and was much more
informative than other quantities measuring information about a data
set.  Consider for example the log-information for a data set,
$I=\log_{10}(n/p)$, which reports the information of a data set by
adjusting its sample size by the number of features.  While this is a
reasonable measure, Figure~\ref{anova} shows that $I$ is not
nearly as effective as $\rho$ in predicting imputation accuracy.  The
figure displays the ANOVA effect sizes for $\rho$ and $I$ from a
linear regression in which log relative imputation error was used as
the response.  In addition to $\rho$ and $I$, dependent variables in
the regression also included the type of RF procedure.  The effect
size was defined as the estimated coefficients for the standardized
values of $\rho$ and $I$.  The two dependent variables $\rho$ and $I$
were standardized to have a mean of zero and variance of one which
makes it possible to directly compare their estimated coefficients.
The figure shows that both values are important for understanding
imputation accuracy and that both exhibit the same pattern.  Within a
specific type of missing data mechanism, say MCAR, importance of each
variable decreases with missingness of data (MCAR 0.25, MCAR 0.5, MCAR
0.75).  However, while the pattern of the two measures is similar, the
effect size of $\rho$ is generally much larger than $I$.  The only
exceptions being the MAR 0.75 and NMAR 0.75 experiments, but these two
experiments are the least interesting.  As will be discussed below,
nearly all methods performed poorly here.

\begin{figure*}[htp!]
\begin{center}
\resizebox{6.0in}{!}{\includegraphics[page=1]{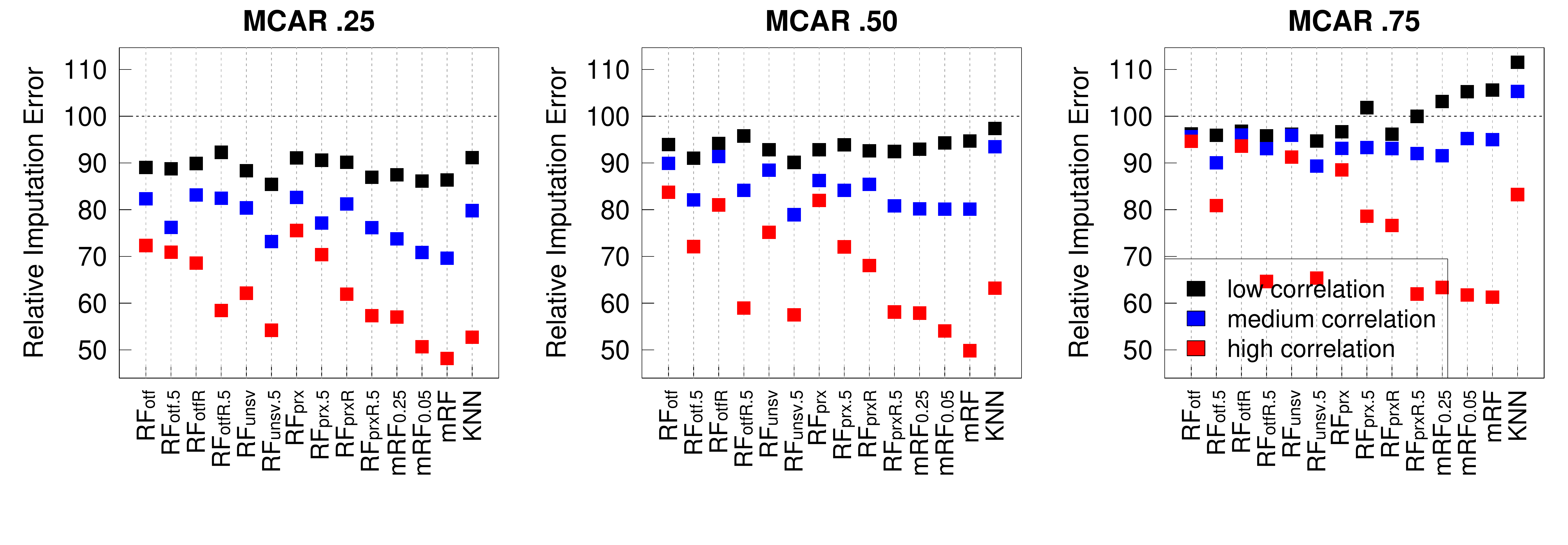}}
\resizebox{6.0in}{!}{\includegraphics[page=2]{benchmark_perf_results.pdf}}
\resizebox{6.0in}{!}{\includegraphics[page=3]{benchmark_perf_results.pdf}}
\end{center}
\vskip-10pt
\caption{Relative imputation error, $\ee_R(\ii)$, stratified and averaged by level of
  correlation of a data set.  Procedures are: \rfo\!, \rfoFive (on the
  fly imputation with 1 and 5 iterations); \rfor\!, \rforFive (similar
  to \rfo and \rfoFive but using pure random splitting); \rfu\!,
  \rfuFive (multivariate unsupervised splitting with 1 and 5
  iterations); \rfp\!, \rfpFive (proximity imputation with 1 and 5
  iterations); \rfpr\!, \rfprFive (same as \rfp and \rfpFive but using
  pure random splitting); \mRF{0.25}, \mRF{0.05}, mRF (mForest
  imputation, with 25\%, 5\% and 1 variable(s) used as the response);
  KNN ($k$-nearest neighbor imputation).}
\label{perf}
\end{figure*}

\begin{table}[phtb!]
\vskip-10pt
\caption{Relative imputation error $\ee_R(\ii)$.}
\footnotesize
\centering
\begin{tabular}{l|rrr|rrr|rrr}
\noalign{\hrule height 1.5pt}
\multicolumn{1}{c}{}& \multicolumn{9}{c}{Low Correlation}\\ 
\multicolumn{1}{c}{}& 
\multicolumn{3}{c}{MCAR}& 
\multicolumn{3}{c}{MAR}& 
\multicolumn{3}{c}{NMAR} \\ 
\multicolumn{1}{c}{}& 
\multicolumn{1}{c}{.25}&
\multicolumn{1}{c}{.50}&
\multicolumn{1}{c}{.75}&
\multicolumn{1}{c}{.25}&
\multicolumn{1}{c}{.50}&
\multicolumn{1}{c}{.75}&
\multicolumn{1}{c}{.25}&
\multicolumn{1}{c}{.50}&
\multicolumn{1}{c}{.75}\\
  \hline
\rfo & 89.0 & 93.9 & 96.2 & 89.5 & 94.5 & 97.2 & 96.5 & 97.2 & 100.9 \\ 
  \rfoFive & 88.7 & 91.0 & 95.9 & 89.5 & 88.6 & 93.5 & 96.0 & 92.6 & 98.8 \\ 
  \rfor & 89.9 & 94.1 & 96.8 & 89.8 & 94.7 & 97.8 & 96.6 & 97.6 & 101.7 \\ 
  \rforFive & 92.3 & 95.8 & 95.8 & 96.5 & 93.7 & 94.2 & 103.2 & 97.0 & 102.9 \\ 
  \rfu & 88.3 & 92.8 & 96.2 & 87.9 & 93.0 & 97.3 & 95.4 & 97.4 & 101.6 \\ 
  \rfuFive & 85.4 & 90.1 & 94.7 & 85.7 & 88.6 & 92.2 & 97.7 & 93.0 & 100.8 \\ 
  \rfp & 91.1 & 92.8 & 96.7 & 89.9 & 88.5 & 90.4 & 91.5 & 92.7 & 99.2 \\ 
  \rfpFive & 90.6 & 93.9 & 101.8 & 89.8 & 88.7 & 93.9 & 95.7 & 91.1 & 99.3 \\ 
  \rfpr & 90.2 & 92.6 & 96.2 & 89.4 & 88.8 & 90.7 & 94.6 & 97.5 & 100.5 \\ 
  \rfprFive & 86.9 & 92.4 & 100.0 & 88.1 & 88.8 & 94.8 & 96.2 & 94.3 & 102.8 \\ 
  \mRF{0.25} & 87.4 & 92.9 & 103.1 & 88.8 & 89.3 & 99.8 & 96.9 & 92.3 & 98.7 \\ 
  \mRF{0.05} & 86.1 & 94.3 & 105.3 & 86.0 & 88.7 & 102.7 & 96.8 & 92.6 & 99.0 \\ 
  mRF & 86.3 & 94.7 & 105.6 & 84.4 & 88.6 & 103.3 & 96.7 & 92.5 & 98.8 \\ 
  KNN & 91.1 & 97.4 & 111.5 & 94.4 & 100.9 & 106.1 & 100.9 & 100.0 & 101.7 \\ 
\multicolumn{9}{c}{}\\   
\multicolumn{1}{c}{}& \multicolumn{9}{c}{Medium Correlation}\\ 
\multicolumn{1}{c}{}& 
\multicolumn{3}{c}{MCAR}& 
\multicolumn{3}{c}{MAR}& 
\multicolumn{3}{c}{NMAR} \\ 
\multicolumn{1}{c}{}& 
\multicolumn{1}{c}{.25}&
\multicolumn{1}{c}{.50}&
\multicolumn{1}{c}{.75}&
\multicolumn{1}{c}{.25}&
\multicolumn{1}{c}{.50}&
\multicolumn{1}{c}{.75}&
\multicolumn{1}{c}{.25}&
\multicolumn{1}{c}{.50}&
\multicolumn{1}{c}{.75}\\
  \hline
\rfo & 82.3 & 89.9 & 95.6 & 78.8 & 88.6 & 97.0 & 92.7 & 92.6 & 102.2 \\ 
  \rfoFive & 76.2 & 82.1 & 90.0 & 83.4 & 79.1 & 93.4 & 99.6 & 89.1 & 100.8 \\ 
  \rfor & 83.1 & 91.4 & 96.0 & 80.3 & 90.3 & 97.4 & 92.2 & 96.1 & 105.3 \\ 
  \rforFive & 82.4 & 84.1 & 93.1 & 88.2 & 84.2 & 95.1 & 112.0 & 97.1 & 104.5 \\ 
  \rfu & 80.4 & 88.4 & 95.9 & 76.1 & 87.7 & 97.5 & 87.3 & 92.7 & 104.7 \\ 
  \rfuFive & 73.2 & 78.9 & 89.3 & 78.8 & 79.0 & 92.4 & 98.8 & 92.8 & 104.2 \\ 
  \rfp & 82.6 & 86.3 & 93.1 & 80.7 & 80.5 & 97.7 & 88.6 & 93.8 & 99.5 \\ 
  \rfpFive & 77.1 & 84.1 & 93.3 & 86.5 & 77.0 & 92.1 & 98.1 & 93.7 & 101.0 \\ 
  \rfpr & 81.2 & 85.4 & 93.1 & 80.4 & 82.4 & 96.3 & 89.2 & 97.2 & 101.3 \\ 
  \rfprFive & 76.1 & 80.8 & 92.0 & 82.1 & 77.7 & 95.1 & 102.1 & 96.6 & 105.1 \\ 
  \mRF{0.25} & 73.8 & 80.2 & 91.6 & 75.3 & 75.6 & 90.2 & 97.6 & 87.5 & 102.1 \\ 
  \mRF{0.05} & 70.9 & 80.1 & 95.2 & 70.1 & 76.6 & 93.0 & 87.4 & 87.9 & 103.4 \\ 
  mRF & 69.6 & 80.1 & 95.0 & 71.3 & 74.6 & 92.4 & 86.9 & 87.8 & 103.1 \\ 
  KNN & 79.8 & 93.5 & 105.3 & 80.2 & 96.0 & 98.7 & 93.9 & 98.3 & 102.1 \\ 
\multicolumn{9}{c}{}\\   
\multicolumn{1}{c}{}& \multicolumn{9}{c}{High Correlation}\\ 
\multicolumn{1}{c}{}& 
\multicolumn{3}{c}{MCAR}& 
\multicolumn{3}{c}{MAR}& 
\multicolumn{3}{c}{NMAR} \\ 
\multicolumn{1}{c}{}& 
\multicolumn{1}{c}{.25}&
\multicolumn{1}{c}{.50}&
\multicolumn{1}{c}{.75}&
\multicolumn{1}{c}{.25}&
\multicolumn{1}{c}{.50}&
\multicolumn{1}{c}{.75}&
\multicolumn{1}{c}{.25}&
\multicolumn{1}{c}{.50}&
\multicolumn{1}{c}{.75}\\
  \hline
\rfo & 72.3 & 83.7 & 94.6 & 65.5 & 83.3 & 98.4 & 66.5 & 84.8 & 100.4 \\ 
  \rfoFive & 70.9 & 72.1 & 80.9 & 69.5 & 70.9 & 91.0 & 70.1 & 70.8 & 97.3 \\ 
  \rfor & 68.6 & 81.0 & 93.6 & 59.5 & 87.1 & 98.9 & 61.2 & 88.2 & 100.3 \\ 
  \rforFive & 58.4 & 58.9 & 64.6 & 56.7 & 55.1 & 88.4 & 58.4 & 60.9 & 97.3 \\ 
  \rfu & 62.1 & 75.1 & 91.3 & 56.8 & 70.8 & 97.8 & 58.1 & 73.3 & 100.6 \\ 
  \rfuFive & 54.2 & 57.5 & 65.4 & 54.0 & 49.4 & 80.0 & 55.4 & 51.7 & 90.7 \\ 
  \rfp & 75.5 & 82.0 & 88.5 & 70.7 & 72.8 & 94.3 & 70.9 & 74.3 & 102.0 \\ 
  \rfpFive & 70.4 & 72.0 & 78.6 & 69.7 & 71.2 & 90.3 & 70.0 & 72.2 & 98.2 \\ 
  \rfpr & 61.9 & 68.1 & 76.6 & 58.7 & 64.1 & 79.5 & 60.4 & 74.6 & 97.5 \\ 
  \rfprFive & 57.3 & 58.1 & 61.9 & 55.9 & 54.1 & 71.9 & 57.8 & 60.2 & 93.7 \\ 
  \mRF{0.25} & 57.0 & 57.9 & 63.3 & 55.5 & 50.4 & 70.5 & 56.7 & 50.7 & 87.3 \\ 
  \mRF{0.05} & 50.7 & 54.0 & 61.7 & 48.3 & 48.4 & 74.9 & 49.9 & 48.6 & 85.9 \\ 
  mRF & 48.2 & 49.8 & 61.3 & 47.0 & 47.5 & 70.2 & 46.6 & 47.6 & 82.9 \\ 
  KNN & 52.7 & 63.2 & 83.2 & 52.0 & 71.1 & 96.4 & 53.2 & 74.9 & 99.2  \\ 
\noalign{\hrule height 1.5pt}
\end{tabular}
\label{perf.table}
\end{table}

\subsubsection{Correlation}

Figure~\ref{perf} and Table~\ref{perf.table}, which have been
stratified by correlation group, show the importance of correlation
for RF imputation procedures.  In general, imputation accuracy
generally improves with correlation.  Over the high correlation data,
mForest algorithms were by far the best.  In some cases, they achieved
a relative imputation error of 50, which means their imputation error
was half of the strawman's value.  Generally there are no noticeable
differences between mRF (missForest) and \mRF{0.05}.  Performance of
\mRF{0.25}, which uses only 4 regressions per cycle (as opposed to
$p$ for mRF), is also very good.  Other algorithms that performed well
in high correlation settings were \rfprFive (proximity imputation with
random splitting, iterated 5 times) and \rfuFive (unsupervised
multivariate splitting, iterated 5 times).  Of these, \rfuFive tended
to perform slightly better in the medium and low correlation settings.
We also note that while mForest also performed well over medium
correlation settings, performance was not superior to other RF
procedures in low correlation settings, and sometimes was worse than
procedures like \rfuFive\!.  Regarding the comparison procedure KNN,
while its performance also improved with increasing correlation,
performance in the medium and low correlation settings was generally
much worse than RF methods.

\subsubsection{Missing data mechanism}

The missing
data mechanism also plays an important role in accuracy of RF procedures.
Accuracy decreased systematically when going from
MCAR to MAR and NMAR.  Except for heavy missingness (75\%), all RF
procedures under MCAR and MAR were more accurate than strawman
imputation.  Performance in NMAR was generally poor unless correlation
was high.

\subsubsection{Heavy missingness}

Accuracy degraded with increasing missingness.
This was especially true when missingness was high (75\%).  For NMAR
data with heavy missingness, procedures were not much better than
strawman (and sometimes worse), regardless of correlation.  However,
even with  missingness of up to 50\%, if correlation was high,
RF procedures could still reduce the strawman's error by one-half.

\subsubsection{Iterating RF algorithms}

Iterating generally improved accuracy for RF algorithms, except in the
case of NMAR data, where in low and medium correlation settings,
performance sometimes degraded.

\subsection{Computational speed}

\begin{figure*}[phtb!]
\begin{center}
\resizebox{1.95in}{!}{\includegraphics[page=1]{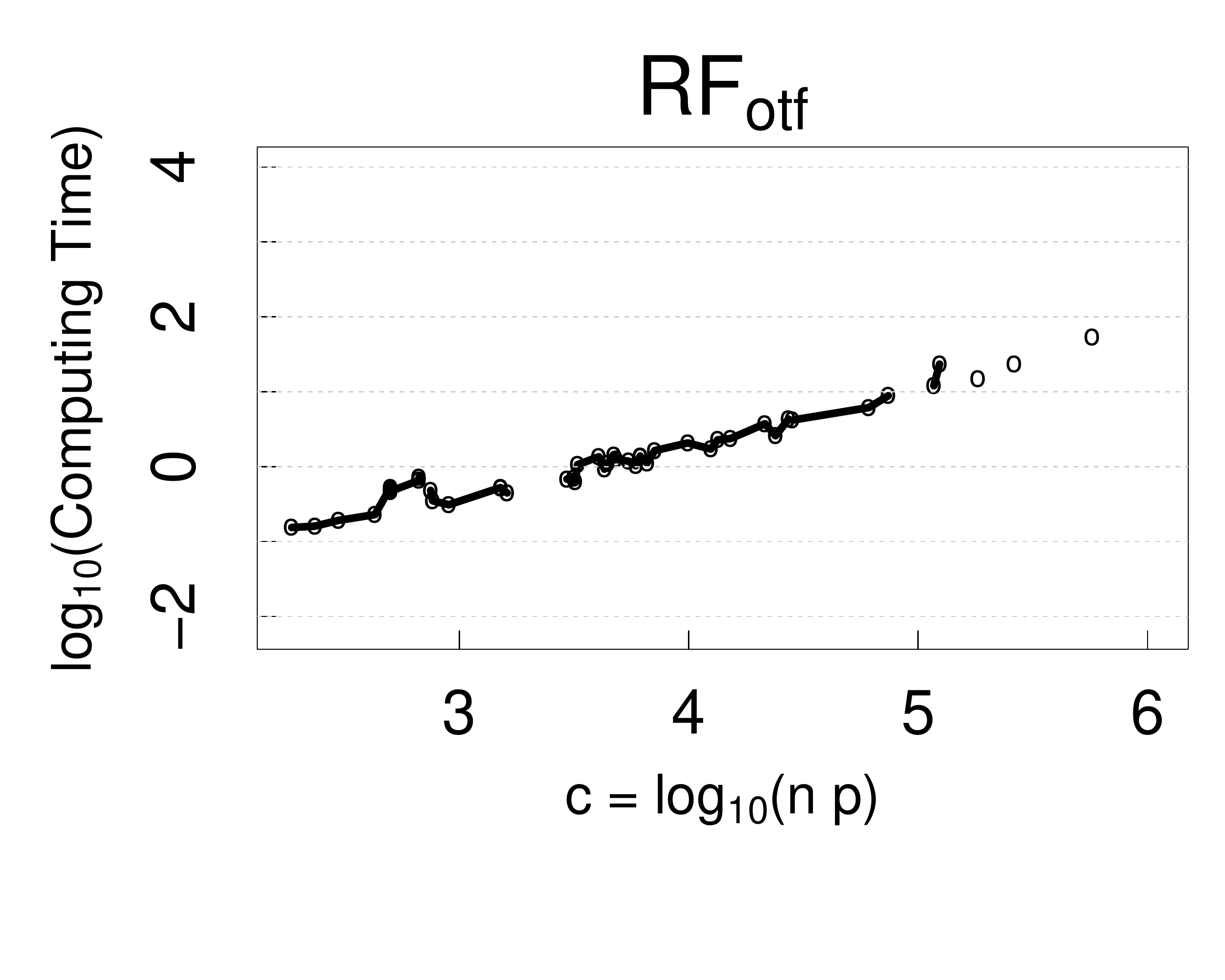}}
\resizebox{1.95in}{!}{\includegraphics[page=2]{benchmark_cpu_results.pdf}}
\resizebox{1.95in}{!}{\includegraphics[page=3]{benchmark_cpu_results.pdf}}
\resizebox{1.95in}{!}{\includegraphics[page=4]{benchmark_cpu_results.pdf}}
\resizebox{1.95in}{!}{\includegraphics[page=5]{benchmark_cpu_results.pdf}}
\resizebox{1.95in}{!}{\includegraphics[page=6]{benchmark_cpu_results.pdf}}
\resizebox{1.95in}{!}{\includegraphics[page=7]{benchmark_cpu_results.pdf}}
\resizebox{1.95in}{!}{\includegraphics[page=8]{benchmark_cpu_results.pdf}}
\resizebox{1.95in}{!}{\includegraphics[page=9]{benchmark_cpu_results.pdf}}
\resizebox{1.95in}{!}{\includegraphics[page=10]{benchmark_cpu_results.pdf}}
\resizebox{1.95in}{!}{\includegraphics[page=11]{benchmark_cpu_results.pdf}}
\resizebox{1.95in}{!}{\includegraphics[page=12]{benchmark_cpu_results.pdf}}
\resizebox{1.95in}{!}{\includegraphics[page=13]{benchmark_cpu_results.pdf}}
\resizebox{1.95in}{!}{\includegraphics[page=14]{benchmark_cpu_results.pdf}}
\end{center}
\vskip-10pt
\caption{Log of computing time for a procedure versus log-computational
  complexity of a data set, $c=\log_{10}(np)$.}
\label{cpu}
\end{figure*}

\begin{figure*}[phtb!]
\begin{center}
\resizebox{1.95in}{!}{\includegraphics[page=15]{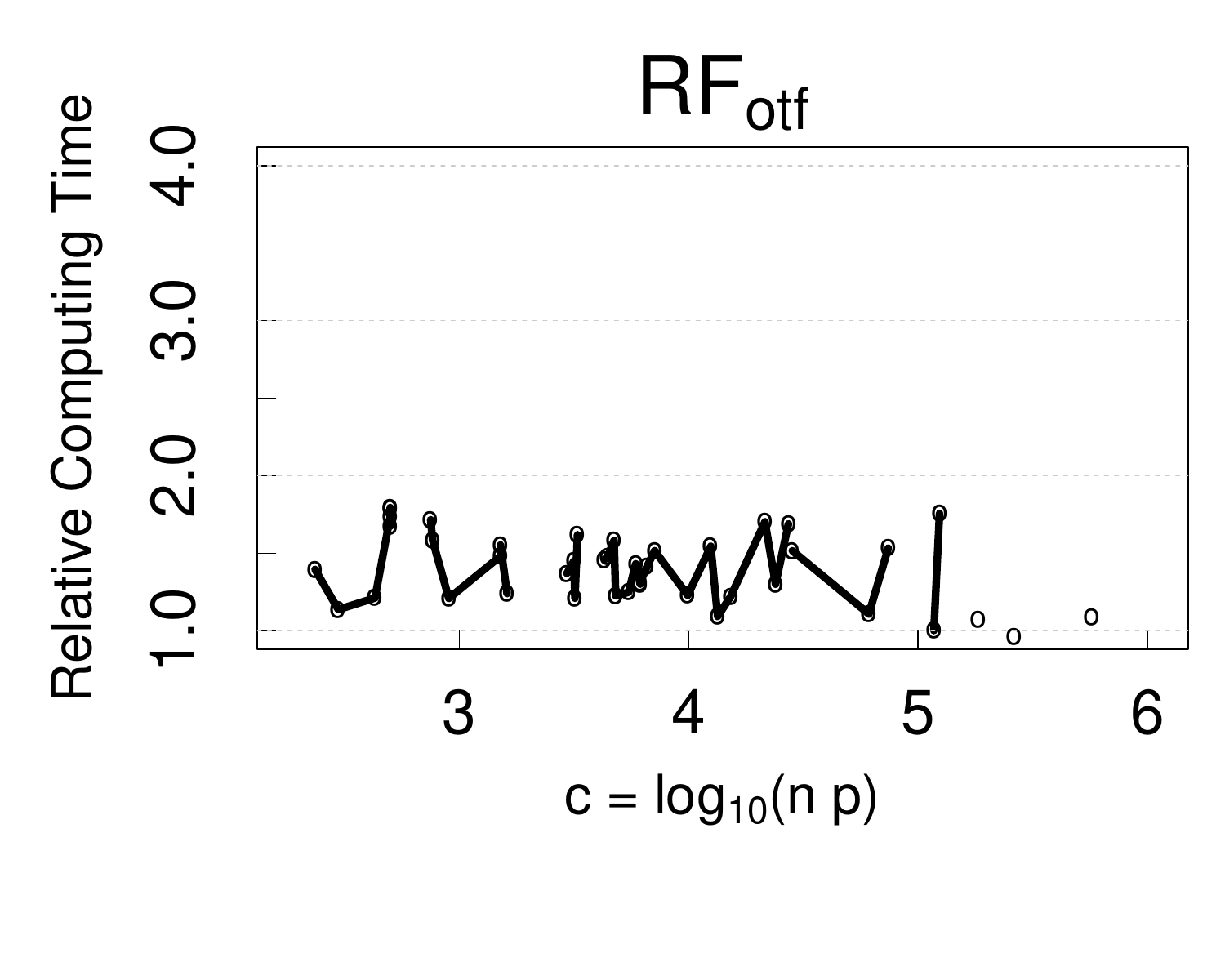}}
\resizebox{1.95in}{!}{\includegraphics[page=16]{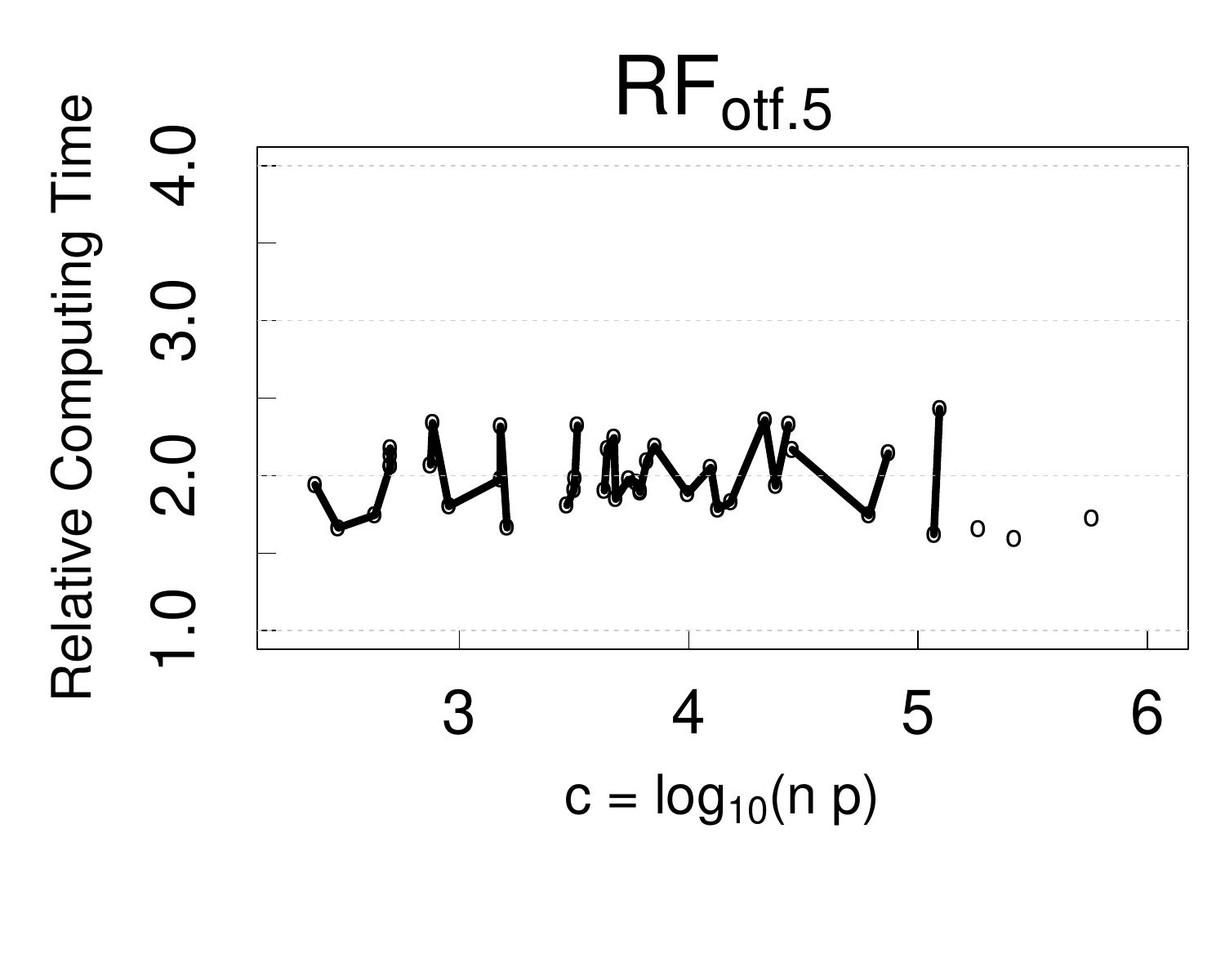}}
\resizebox{1.95in}{!}{\includegraphics[page=17]{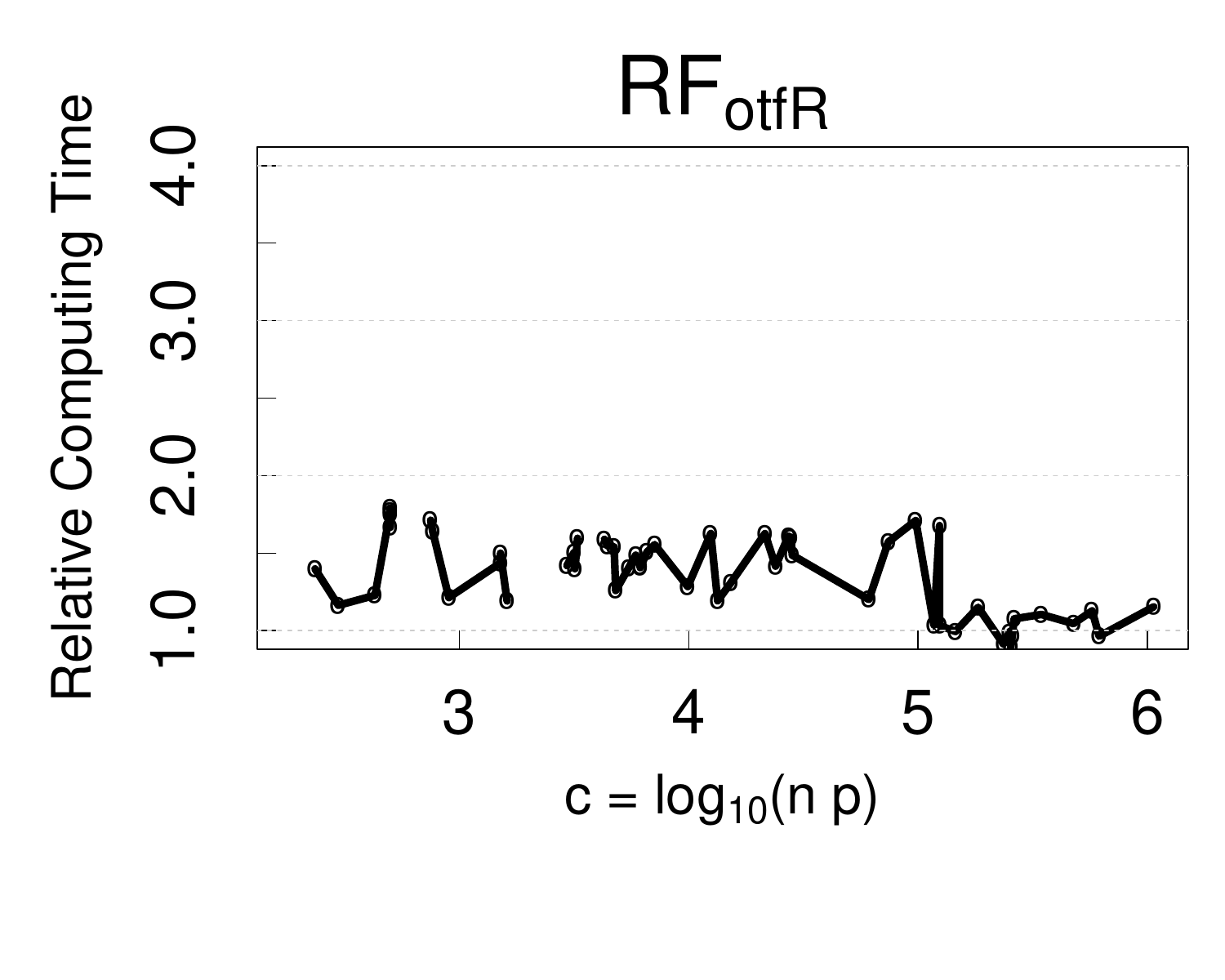}}
\resizebox{1.95in}{!}{\includegraphics[page=18]{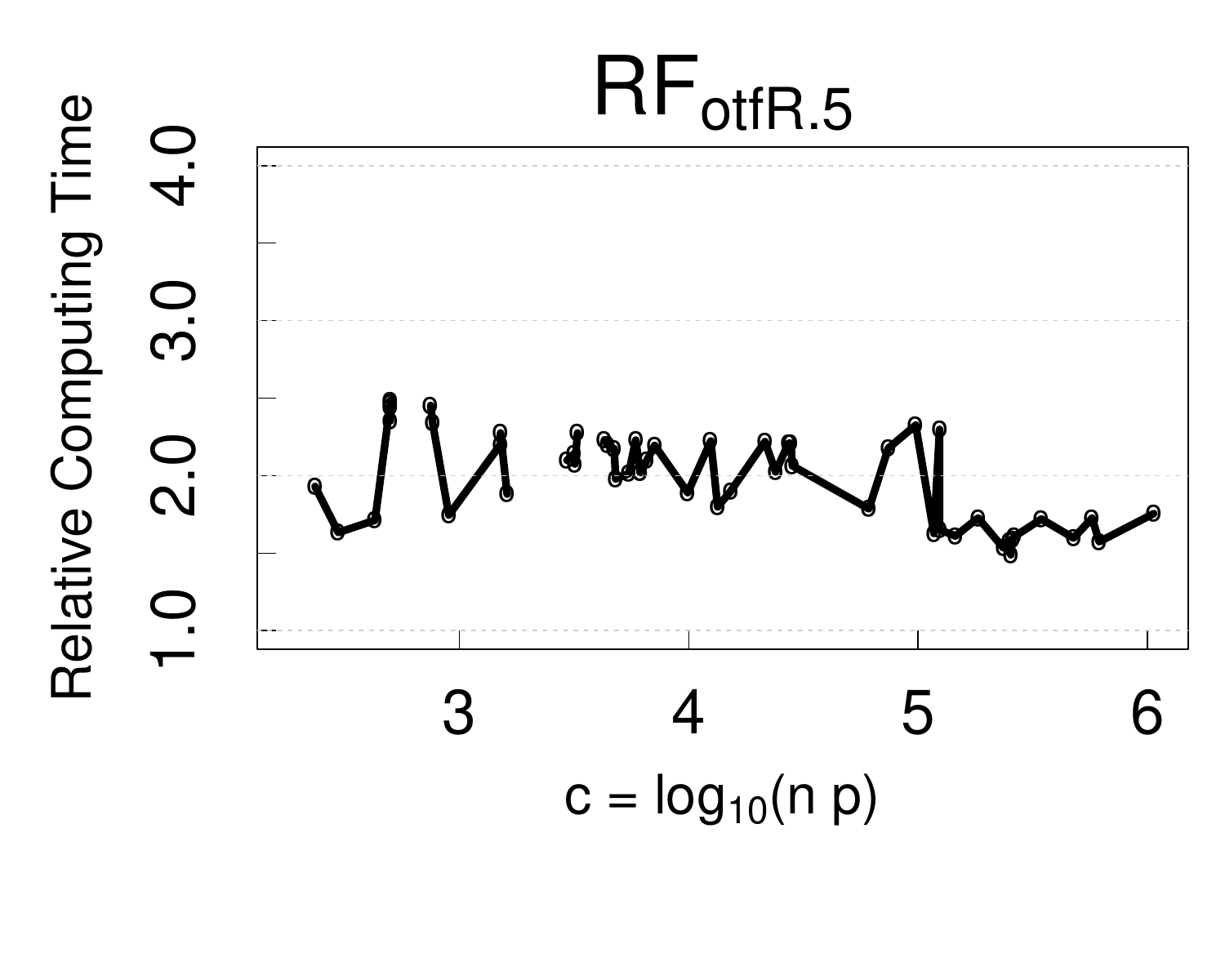}}
\resizebox{1.95in}{!}{\includegraphics[page=19]{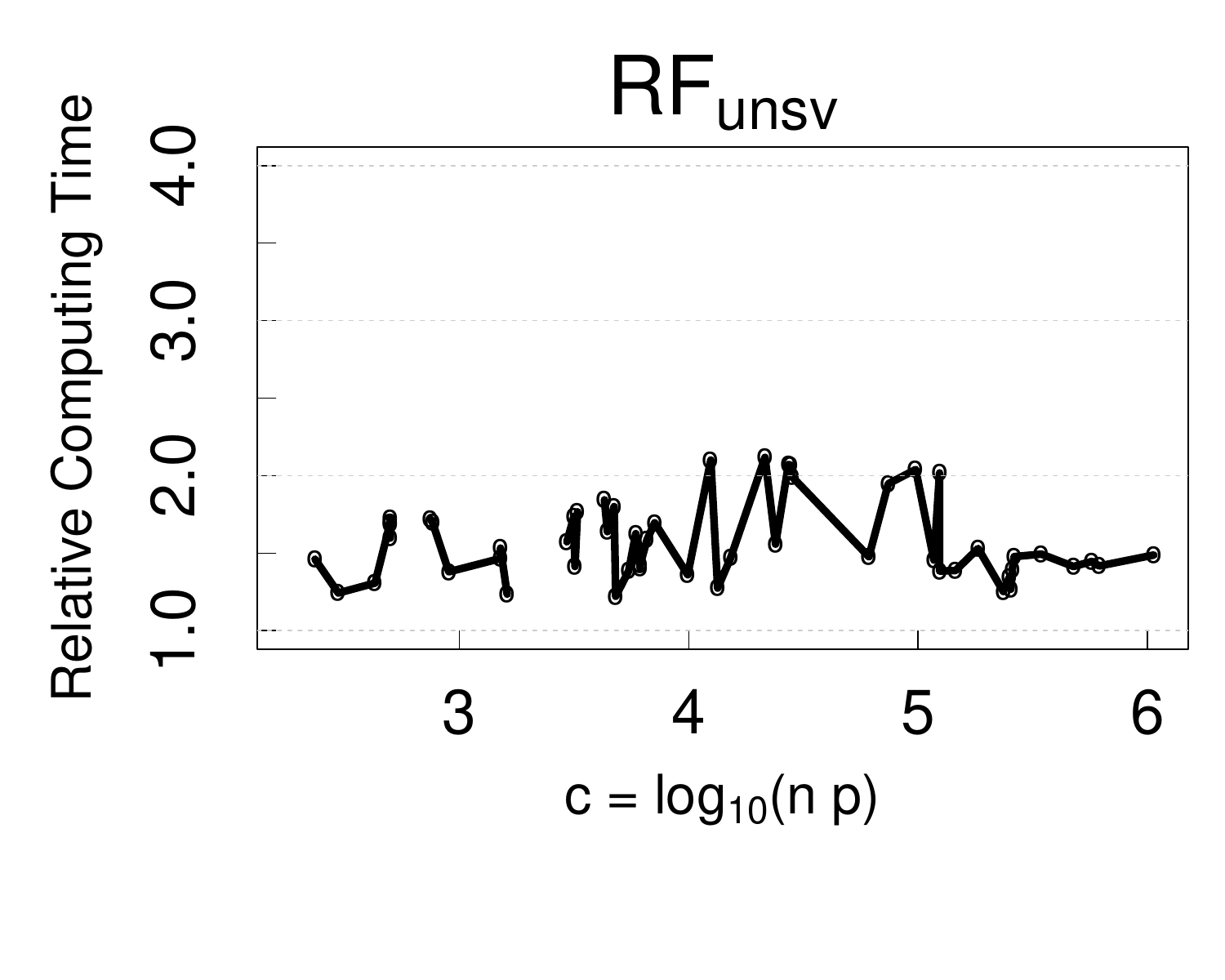}}
\resizebox{1.95in}{!}{\includegraphics[page=20]{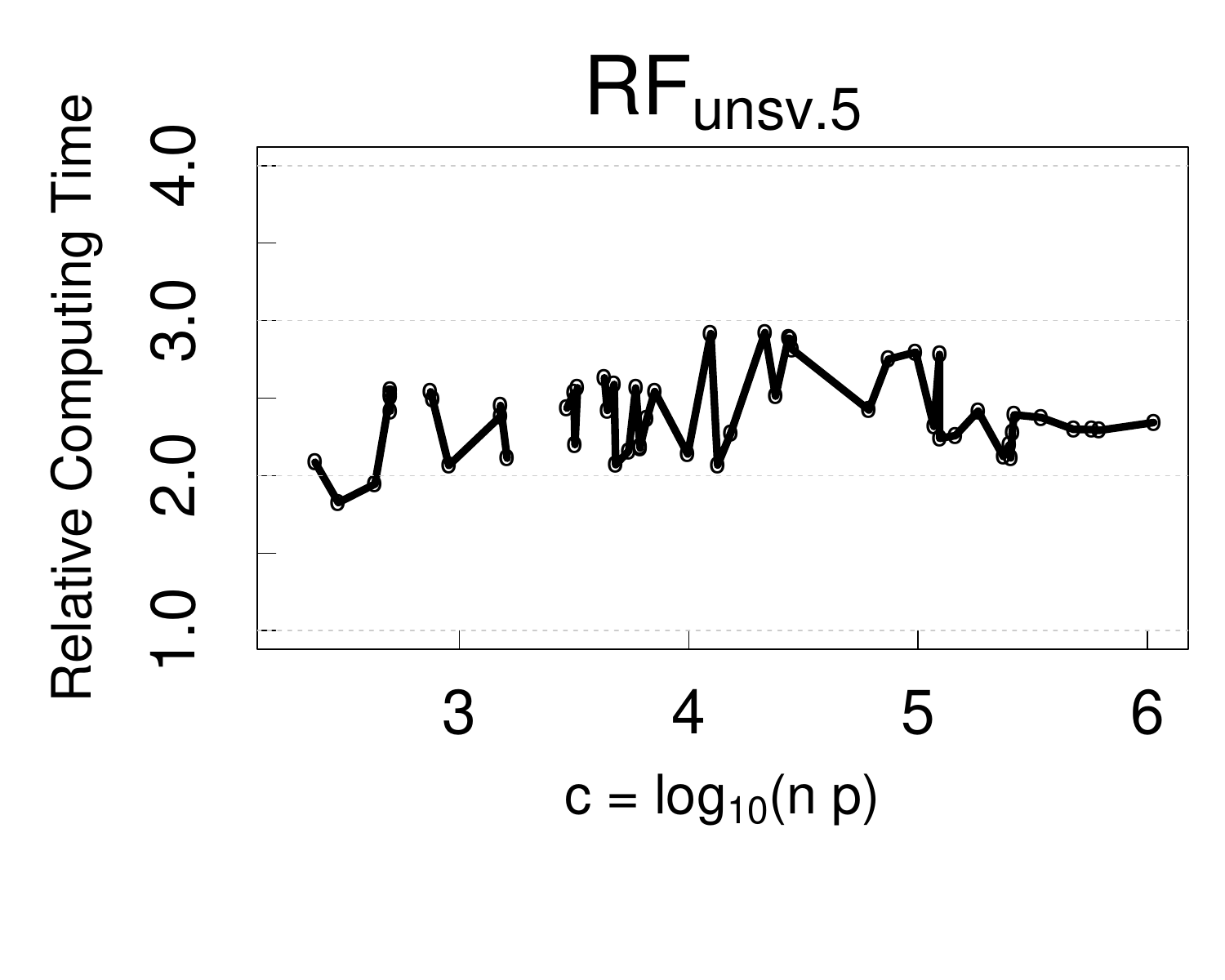}}
\resizebox{1.95in}{!}{\includegraphics[page=21]{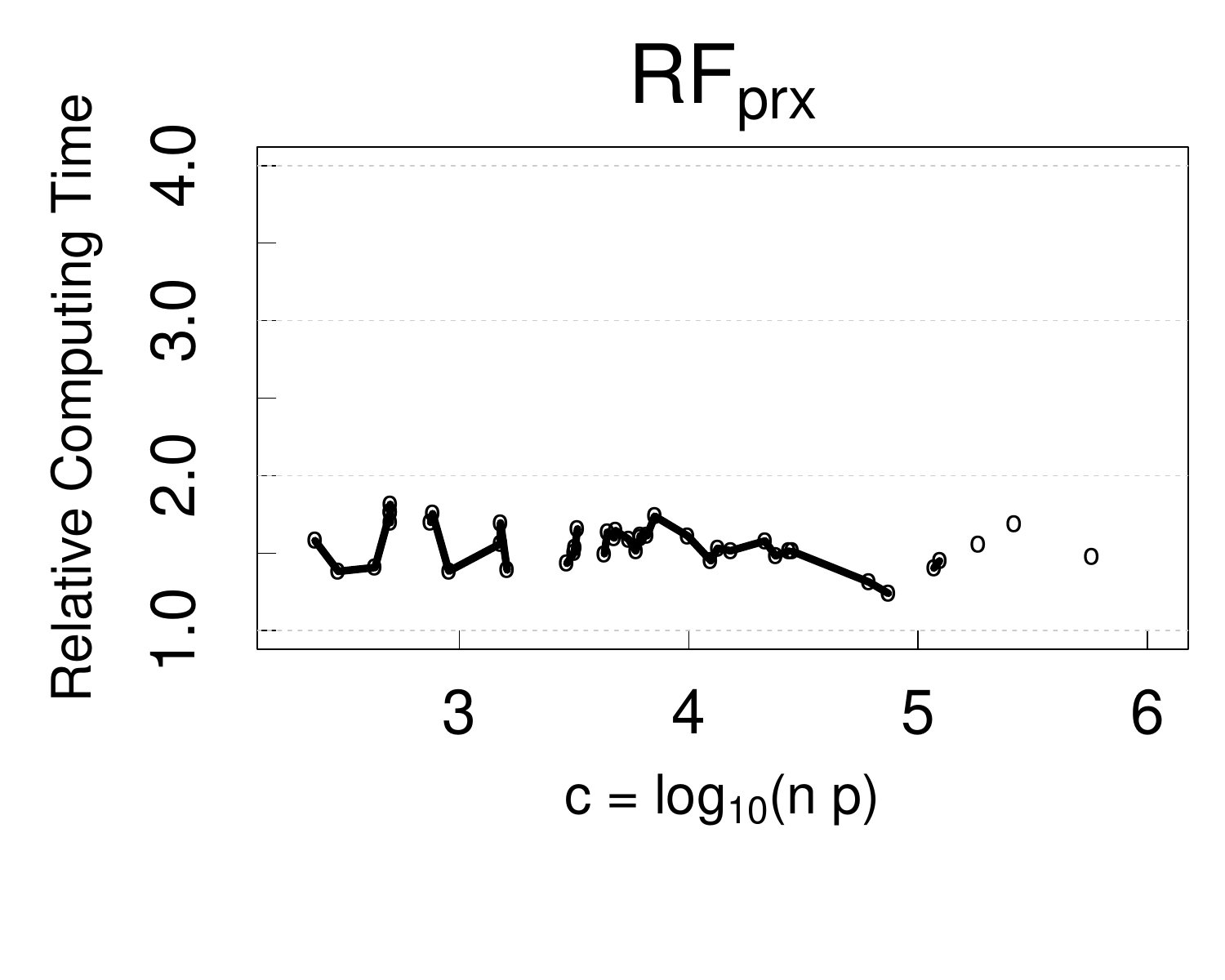}}
\resizebox{1.95in}{!}{\includegraphics[page=22]{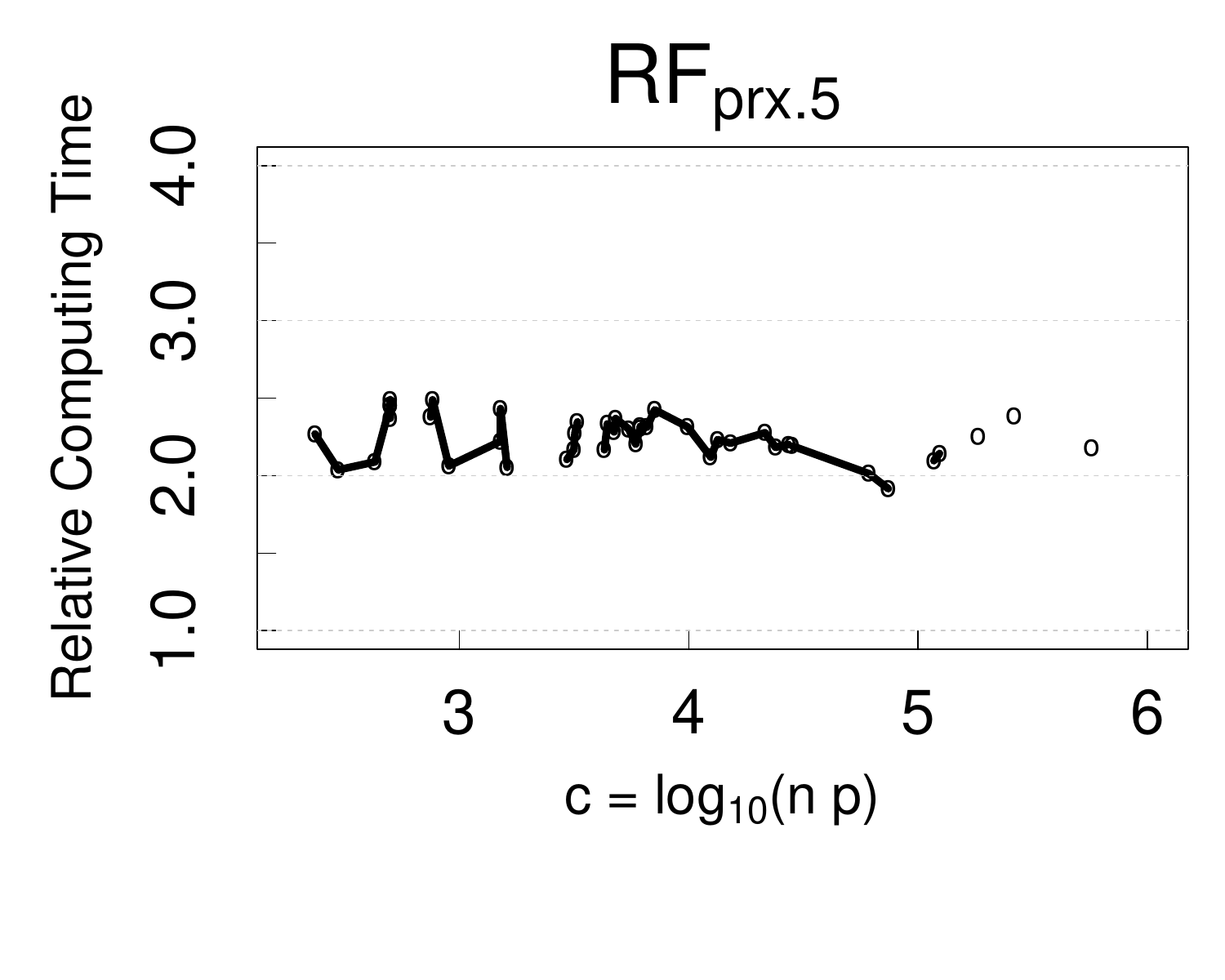}}
\resizebox{1.95in}{!}{\includegraphics[page=23]{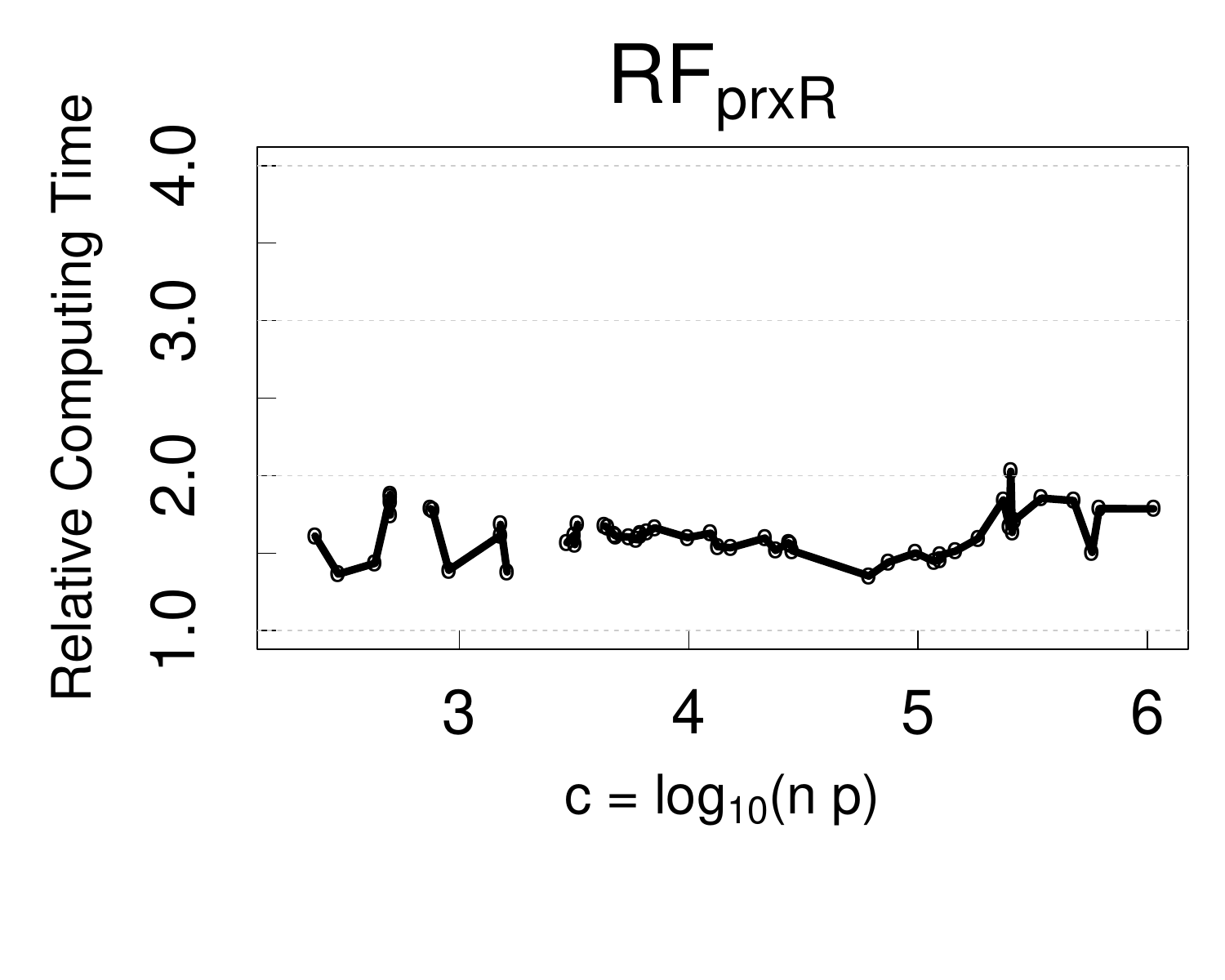}}
\resizebox{1.95in}{!}{\includegraphics[page=24]{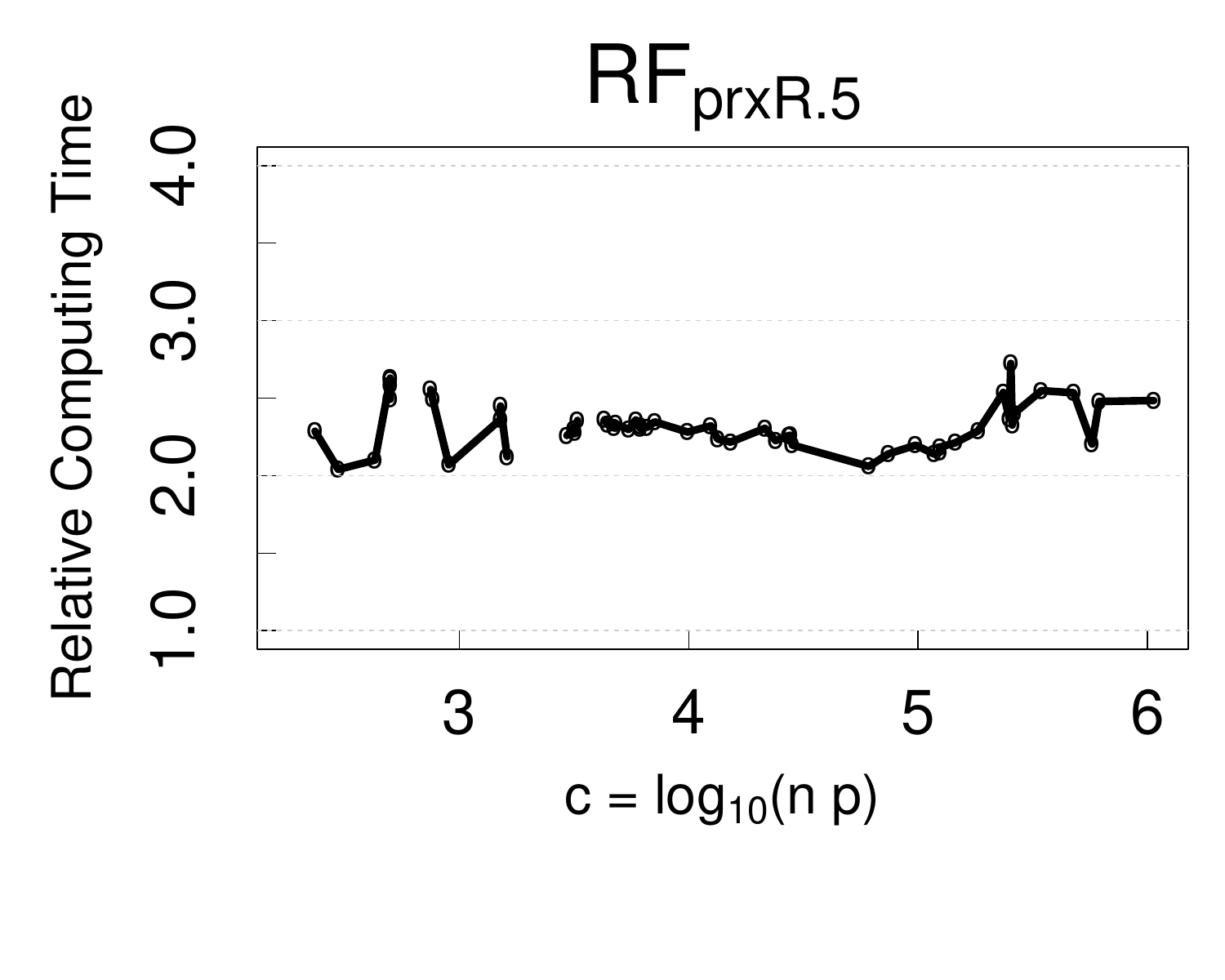}}
\resizebox{1.95in}{!}{\includegraphics[page=25]{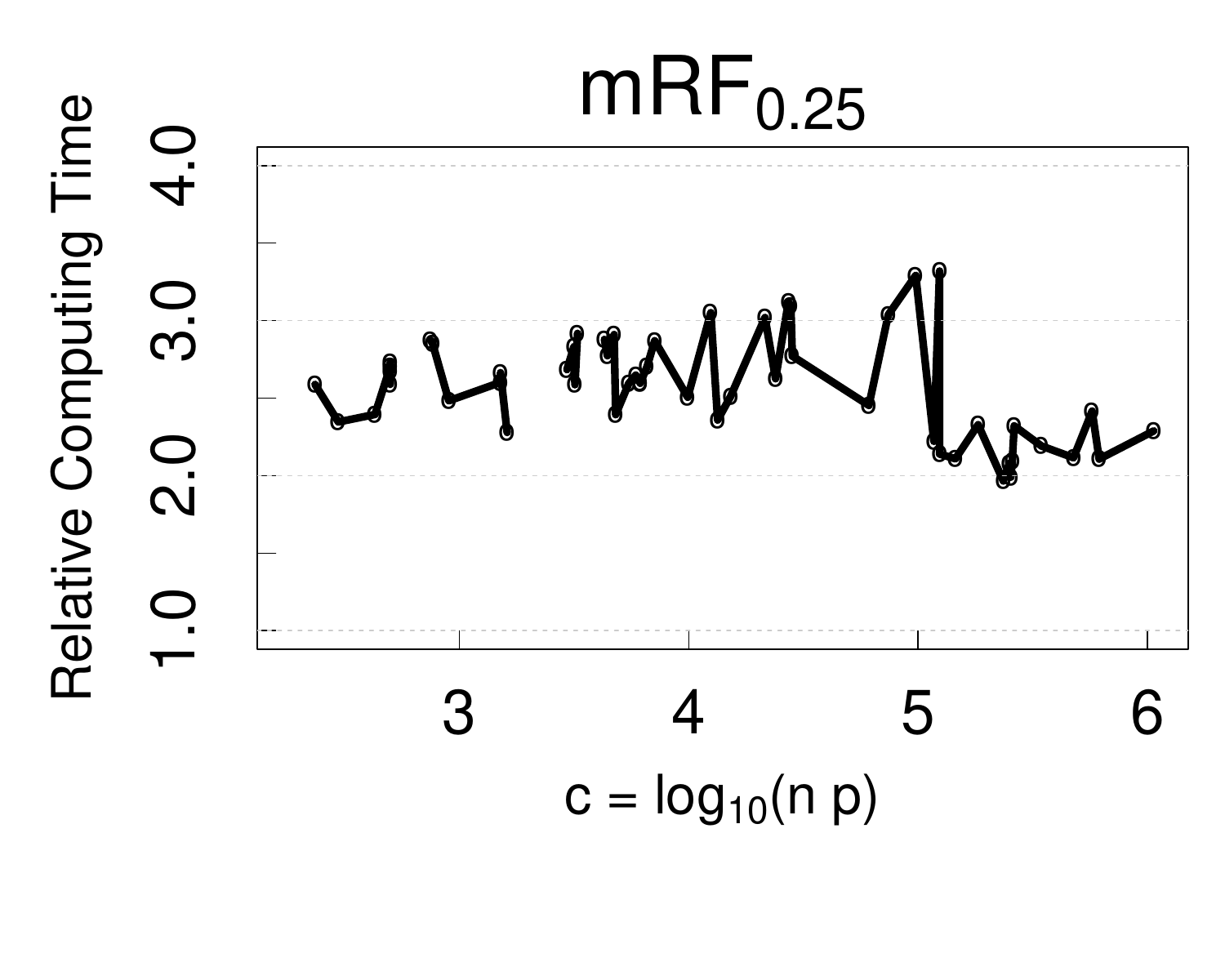}}
\resizebox{1.95in}{!}{\includegraphics[page=26]{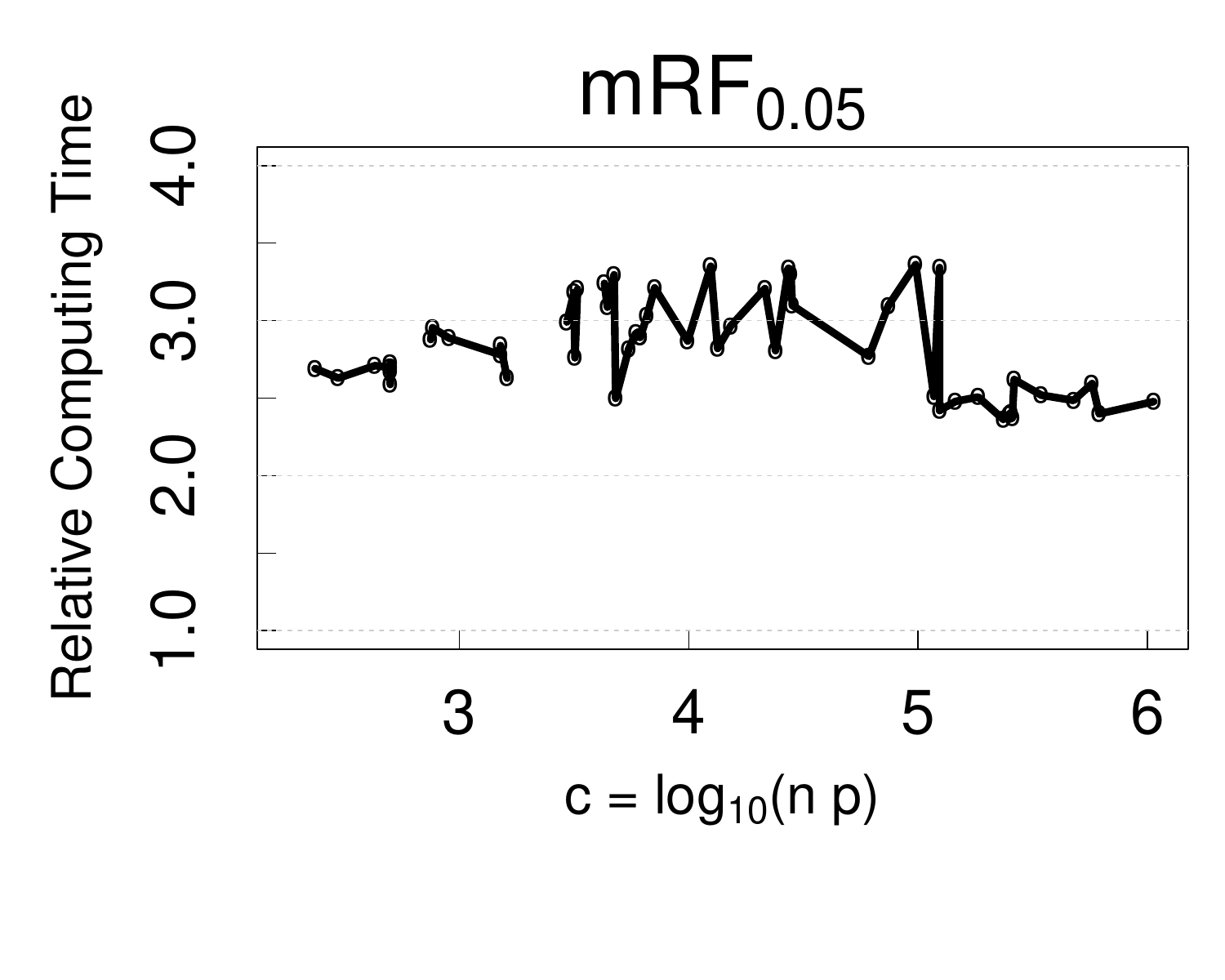}}
\resizebox{1.95in}{!}{\includegraphics[page=27]{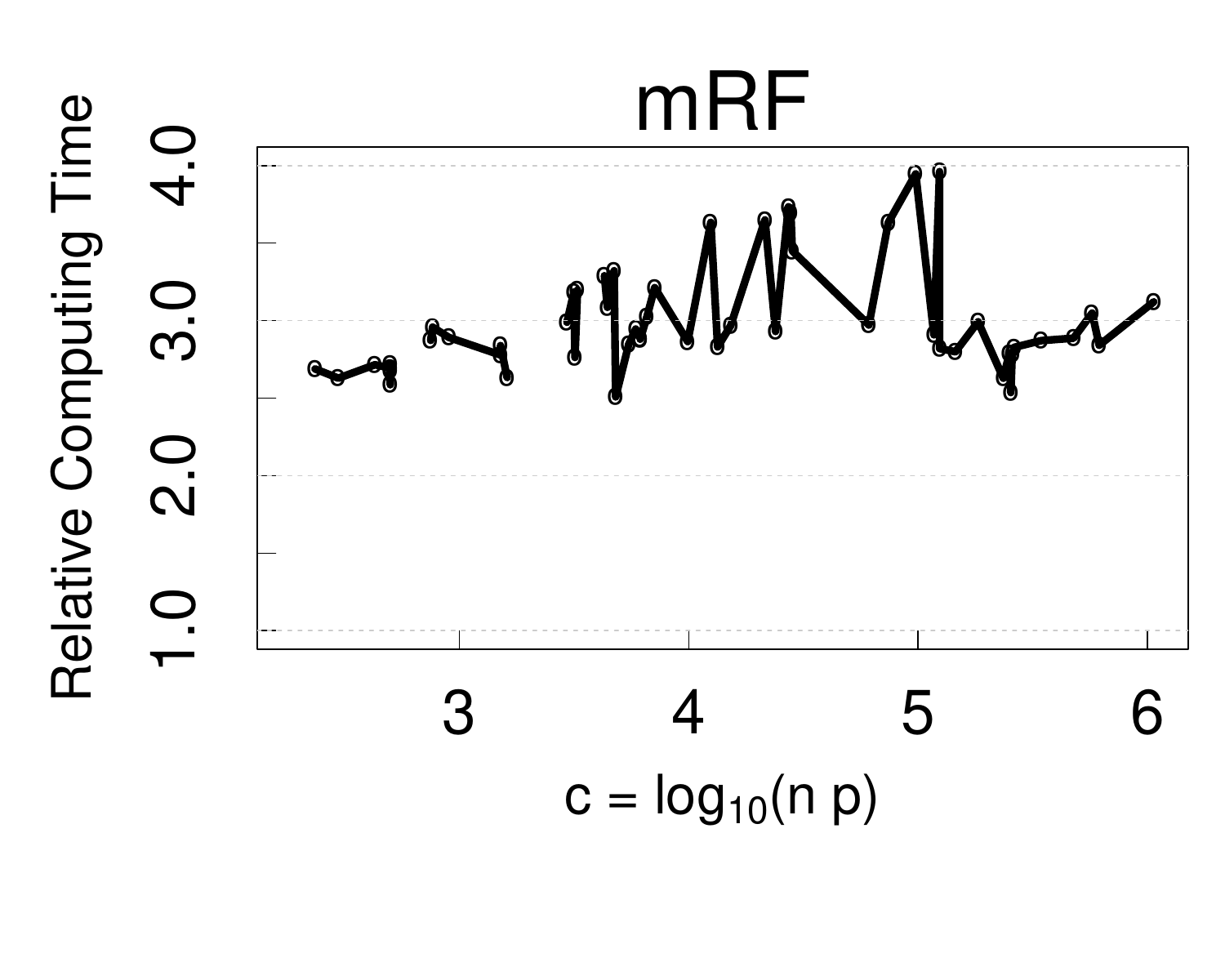}}
\end{center}
\vskip-10pt
\caption{Relative log-computing time (relative to KNN)
versus log-computational
  complexity of a data set, $c=\log_{10}(np)$.}
\label{cpu.relative}
\end{figure*}

Figure~\ref{cpu} displays the log of total elapsed time of a procedure
averaged over all experimental conditions and runs, with results
ordered by the log-computational complexity of a data set,
$c=\log_{10}(np)$.  The fastest algorithm is KNN which is generally 3
times faster on the log-scale, or
1000 times faster, than the slowest algorithm, mRF (missForest).  To
improve clarity of these differences, Figure~\ref{cpu.relative}
displays the relative computational time of procedure relative to KNN
(obtained by subtracting the KNN log-time from each procedures log-time).
This new figure shows that while mRF is 1000 times slower than KNN,
the multivariate mForest algorithms, \mRF{0.05} and \mRF{0.25},
improve speeds by about a factor of 10.  After this, the next slowest
procedures are the iterated algorithms.  Following this are the
non-iterated algorithms.  Some of these latter algorithms, such as
\rfo\!, are 100 times faster than missForest; or only 10 times slower
than KNN.  These kinds of differences can have a real effect when
dealing with big data.  We have experienced settings where OTF
algorithms can take hours to run.  This means that the same data would
take missForest 100's of hours to run, which makes it questionable to
be used in such settings.

\section{Simulations}

In this section we used simulations to study the peformance of
RF as the sample size $n$ was varied.
We wanted to verify two questions: (1) Does the
relative imputation error improve with sample size?
(2) Do these values converge to the same or different values for the
different RF imputation algorithms?

For our simulations, there were 10 variables $X_1,\ldots,X_{10}$ where
the true model was
$$
Y= X_1 + X_2 + X_3 + X_4 + \varepsilon
$$ where $\varepsilon$ was simulated independently from a N$(0,0.5)$
distribution.  Variables $X_1$ and $X_2$ were correlated with a
correlation coefficient of 0.96, and $X_5$ and $X_6$ were correlated
with value 0.96.  The remaining variables were not correlated.
Variables $X_1,X_2,X_5,X_6$ were N$(3,3)$, variables $X_3,X_{10}$ were
N$(1,1)$, variable $X_8$ was N$(3,4)$, and variables $X_4,X_7,X_9$ were
exponentially distributed with mean 0.5.

\begin{figure*}[htp!]
\begin{center}
\resizebox{6.0in}{!}{\includegraphics[page=1]{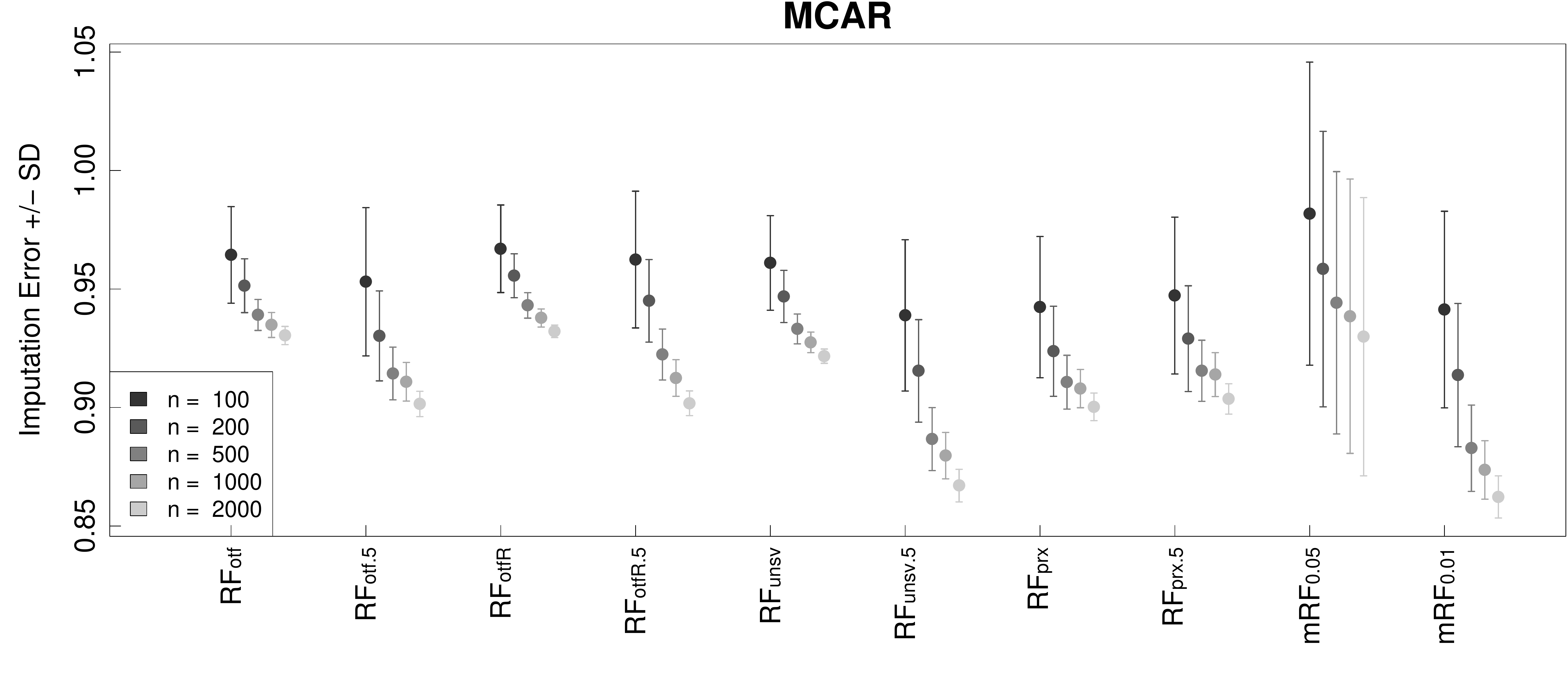}}
\resizebox{6.0in}{!}{\includegraphics[page=2]{simulations.pdf}}
\resizebox{6.0in}{!}{\includegraphics[page=3]{simulations.pdf}}
\end{center}
\vskip-15pt
\caption{Mean relative imputation error $\pm$ standard deviation from
  simulations under different sample size values $n=100,200,500,1000,2000$.}
\label{simulations}
\end{figure*}

The sample size ($n$) was chosen to be 100, 200, 500, 1000, and 2000.
Data was made missing using the MCAR, MAR, and NMAR missing data
procedures described earlier.  Percentage of missing data was set at
25\%. All imputation parameters were set to the same values as used in
our previous experiments as described in Section 3.4.  Each experiment
was repeated 500 times and the relative imputation error, $\ee_R(\ii)$,
recorded in each instance.  Figure~\ref{simulations} displays
the mean relative imputation error for a RF procedure and its standard
deviation for each sample size setting.  As can be seen,
values improve with increasing $n$.  It is also noticeable that
performance depends upon the RF imputation method.  In these
simulations, the missForest algorithm \mRF{0.1} appears to be best
(note that $p=10$ so \mRF{0.1} corresponds to the limiting case missForest).
Also, it should be noted that performance of RF procedures decrease
systematically as the missing data mechanism becomes more complex.
This mirrors our previous findings.

\section{Conclusions}

Being able to deal with missing data effectively is of great importance
to scientists working with real world data today.  A machine learning
method such as RF known for its excellent prediction peformance and
ability to handle all forms of data, represents a poentially
attractive solution to this challenging problem.  However, because no
systematic study of RF procedures had been attempted in missing data
settings, we undertook a large scale experimental study of various RF
procedures to determine which methods performed best, and under what
types of settings.  What we found was that correlation played a very
strong role in performance of RF procedures.  Imputation performance
of all RF procedures improved with increasing correlation of features.
This held even with heavy levels of missing data and in all but the
most complex missing data scenarios.  When there is high correlation
we recommend using a method like missForest which performed the best
in such settings.  Although it might seem obvious that increasing
feature correlation should improve imputation, we found that in low to
medium correlation, RF algorithms did noticeably better than the
popular KNN imputation method.  This is interesting because KNN is
related to RF.  Both methods are a type of nearest neighbor method,
although RF is more adaptive than KNN and in fact can be more
accurately described as an adaptive nearest neighbor method.  This
adaptivity of RF may play a special role in harnessing correlation in
the data that may not necessarily be present in other methods, even
methods that have similarity to RF.  Thus, we feel it is worth
emphasizing that correlation is extremely important to RF
imputation methods.

In big data settings, computational speed will play a key role.  Thus
practically speaking users might not be able to implement the best
method possible because computational times will simply be too long.
This is the downside of a method like missForest, which was the
slowest of all the procedures considered.  As one solution, we
proposed mForest (\mRF{$\a$}) which is a
omputationally more efficient implementation of missForest.
Our results showed mForest could achieve up to
a 10-fold reduction in compute time relative to missForest.  We believe
these computational times can be improved further by incorporating
mForest directly into the native C-library of {\ttfamily
  randomForestSRC} (RF-SRC).  Currently mForest is run as an external
R-loop that makes repeated calls to the {\ttfamily impute.rfsrc}
function in RF-SRC.  Incorporating mForest into the native library,
combined with the openMP parallel processing of RF-SRC, could make it
much more attractive.  However, even with all of this, we still
recommend some of the more basic OTFI algorithms like unsupervised RF
imputation procedures for big data.  These algorithms perform solidly
in terms of imputation and are 100's of times faster than missForest.

\section{Acknowledgement}

This work was supported by the National Institutes of Health [R01CA163739 to H.I.].

\bibliographystyle{natbib}

\end{document}